%% file: main.tex
\documentclass[10pt,journal,compsoc]{IEEEtran}
\ifCLASSOPTIONcompsoc
  \usepackage[nocompress,noadjust]{cite}
\else
  \usepackage{cite}
\fi
\ifCLASSINFOpdf
\else
\fi
\usepackage{epsfig}		
\usepackage{graphicx}	
\usepackage{amsmath}		
\usepackage{amssymb}	
\usepackage{multirow}
\usepackage[linesnumbered,ruled]{algorithm2e}		
\usepackage{comment}		
\newcommand{\etal}{\textit{et al}.~}
\usepackage{booktabs}
\usepackage{xcolor}
\usepackage{xspace}
\usepackage{enumitem}

\newcommand{\logits}{pre-softmax~output\xspace}
\usepackage[hidelinks,citecolor=blue]{hyperref}	

\begin{document}
%
\title{Regularizers for Single-step Adversarial Training}
%
%
%
%
\author{B.S. Vivek, R. Venkatesh Babu,~\IEEEmembership{Senior~Member,~IEEE}
\thanks{The authors are with the Video Analytics Lab, Department of Computational and Data Sciences, Indian Institute of Science, Bangalore, India.\protect\\
E-mail: svivek@iisc.ac.in and venky@iisc.ac.in\protect\\
}
}

%
%

\markboth{}%
{Shell \MakeLowercase{\textit{et al.}}: Bare Demo of IEEEtran.cls for Computer Society Journals}
%



\IEEEtitleabstractindextext{%
\begin{abstract}
The progress in the last decade has enabled machine learning models to achieve impressive performance across a wide range of tasks in Computer Vision. However, a plethora of works have demonstrated the susceptibility of these models to adversarial samples. Adversarial training procedure has been proposed to defend against such adversarial attacks. Adversarial training methods augment mini-batches with adversarial samples, and typically single-step (non-iterative) methods are used for generating these adversarial samples. However, models trained using single-step adversarial training converge to degenerative minima where the model merely appears to be robust. The pseudo robustness of these models is due to the gradient masking effect. Although multi-step adversarial training helps to learn robust models, they are hard to scale due to the use of iterative methods for generating adversarial samples. To address these issues, we propose three different types of regularizers that help to learn robust models using single-step adversarial training methods. The proposed regularizers mitigate the effect of gradient masking by harnessing on properties that differentiate a robust model from that of a pseudo robust model. Performance of models trained using the proposed regularizers is on par with models trained using computationally expensive multi-step adversarial training methods. 
\end{abstract}

}

\maketitle

\IEEEdisplaynontitleabstractindextext

%
\IEEEpeerreviewmaketitle

\IEEEraisesectionheading{\section{Introduction}\label{sec:introduction}}
%
%
%
%
\IEEEPARstart{D}{eep} Neural Networks (DNNs) achieve impressive performance on various tasks in Computer Vision. However, the susceptibility of these networks to adversarial samples~\cite{intriguing-iclr-2014} (samples with crafted noise that can manipulate the model's output) is an important issue. Further, Szegedy~\etal\cite{intriguing-iclr-2014} showed that these adversarial samples are transferable across models of the same or different architectures, and this property enables an attacker to launch attacks on the deployed models in a black-box setting (\cite{papernot2017practical,delving-iclr-2017,dong2018boosting}): where partial or no knowledge of the deployed model is available to the attacker. These properties of adversarial samples pose challenges for the deployment of DNNs in the real world. A plethora of works have proposed various methods to defend against adversarial attacks, such as input transformations (\cite{dziugaite2016study,guo2018countering}), adversarial training (\cite{explainingharnessing-iclr-2015,atscale-iclr-2017,madry-iclr-2018,ensembleAT-iclr-2018}), detection (\cite{feinman2017detecting,xu2017feature}), etc. In this direction, adversarial training method shows promising results, where mini-batches are augmented with adversarial samples, and typically these samples are generated by the model being trained. Further, adversarial samples can be generated by non-iterative (\cite{explainingharnessing-iclr-2015,atscale-iclr-2017})  or iterative methods (\cite{deepfool-cvpr-2016,madry-iclr-2018}). In order to scale adversarial training to large datasets, non-iterative methods such as Fast Gradient Sign Method (FGSM)~\cite{explainingharnessing-iclr-2015} are used. However, models trained using single-step adversarial training methods  (adversarial samples are generated using non-iterative methods) are susceptible to iterative attacks in a white-box setting~\cite{atscale-iclr-2017} (complete knowledge of the deployed model is available to the attacker), and to non-iterative and iterative attacks in a black-box setting (\cite{ensembleAT-iclr-2018,dong2018boosting}).

Tramer~\etal\cite{ensembleAT-iclr-2018} demonstrated that models trained using a single-step adversarial training method converge to  degenerative minima, and this causes gradient masking i.e., the linear approximation of loss becomes unreliable for generating adversarial samples using a non-iterative method. Non-iterative methods such as FGSM generate adversarial perturbations based on the first-order approximation of the loss function i.e., perturbation is in the direction of the sign of the gradient of the loss with respect to the input image. Further, it is implicitly assumed that the model's loss increases for a large perturbation in this direction. This assumption is valid for normally trained models and is not valid for models trained using single-step adversarial training. Madry~\etal\cite{madry-iclr-2018} showed that it is possible to learn robust models, by including adversarial perturbations that maximize the loss while training, and further show that the maximization of loss can be achieved by generating adversarial samples using iterative methods. Unlike non-iterative methods, iterative methods perturb image slightly at each step and this prevents models from exhibiting gradient masking. Though iterative methods help to learn robust models, they are computationally expensive and cause training time to increase drastically. In this work, we propose three different types of regularizers that help to learn robust models using single-step adversarial training methods. The proposed regularizers harness the salient properties of models trained using iterative methods, such as loss surface smoothness and loss monotonicity, and incorporate these properties into models trained using single-step methods. Following are the major contributions of this work:
\begin{itemize}
    \item We bring out the salient properties that differentiate a robust model from that of a pseudo robust model such as loss monotonicity.
    \item Harnessing on the above properties, we propose three different types of regularizers to learn robust models using single-step adversarial methods. The resultant models are robust against both non-iterative and iterative attacks, and achieve on par results when compared to models trained using computationally expensive multi-step adversarial training methods.
\end{itemize}
The paper is organised as follows: section~\ref{sec:notations} introduces the notation followed in this paper, section~\ref{sec:related_works} discusses the related works, section~\ref{sec:proposed_approach}  presents the proposed approach, section~\ref{sec:experiments} hosts the experiments, and section~\ref{sec:discussion_conclusion} concludes the paper.
\section{Notations}\label{sec:notations}
In this section we define the notations followed throughout this paper:
\begin{itemize}[leftmargin=*]
\item $x:$ clean image from the dataset.
\item $y_{true}:$ ground truth label corresponding to the  image $x$.
\item $f(\cdot):$ neural network that maps input image $x$ to the class probabilities.
\item $g(\cdot):$ pre-softmax output of the neural network.
\item $\theta:$ parameters of the neural network.
\item $J:$ loss function e.g., cross-entropy loss.
\item $\nabla_{x}J:$ gradient of the loss with respect to the input image $x$
\item $m:$ mini-batch size.
\item $\epsilon:$ perturbation size of the crafted noise. 
\item $x_{fgsm}:$ potential adversarial sample corresponding to the image $x$, generated using FGSM~\cite{explainingharnessing-iclr-2015}.
\item $x_{ifgsm}:$ potential adversarial sample corresponding to the image $x$, generated using IFGSM~\cite{physicalworld-arxiv-2016}.
\item $x_{rfgsm}:$ potential adversarial sample corresponding to the image $x$, generated using RFGSM~\cite{ensembleAT-iclr-2018}.
\end{itemize}
\section{Related works}\label{sec:related_works}
Following the findings of Szegedy~\etal\cite{intriguing-iclr-2014}, various image specific (e.g.~\cite{explainingharnessing-iclr-2015,physicalworld-arxiv-2016,robustness-arxiv-2016,dong2018boosting}) and image agnostic (e.g.~\cite{universal-cvpr-2017,gduap-mopuri-2018}) attacks have been proposed. Various defense methods (\cite{defensivedistillation-arxiv-2015,explainingharnessing-iclr-2015,buckman2018thermometer,ma2018characterizing,guo2018countering,s.2018stochastic,xie2018mitigating,song2018pixeldefend,samangouei2018defensegan,kannan2018adversarial,S_2019_CVPR_Workshops}) have been proposed to defend against adversarial attacks. In this direction, adversarial training approach~\cite{explainingharnessing-iclr-2015} shows promising results. Kurakin~\etal\cite{atscale-iclr-2017} observed that models trained using single-step adversarial training methods were susceptible to multi-step adversarial attacks in a white-box setting. Further, Tramer~\etal\cite{ensembleAT-iclr-2018} 
demonstrated that the pseudo robustness of these models is due to the gradient masking. Gradient masking causes the first-order approximation of the loss function to become unreliable for generating adversarial samples using non-iterative methods, and this results in the exclusion of useful adversarial samples during training.\\
Madry~\etal\cite{madry-iclr-2018} demonstrated that it is possible to learn models that are robust to single-step and multi-step attacks, if perturbations crafted while training maximize the loss. An iterative method named Projected Gradient Descent (PGD) is used to generate such adversarial samples. Further, Zhang~\etal\cite{Zhang2019theoretically} proposed a regularizer for multi-step adversarial training to encourage the output of the classifier to be smooth. Other line of works such as \cite{wong2017provable,raghunathan2018certified} provide defense certification for norm-bound attacks. However such methods are hard to scale for large datasets and attack perturbation sizes. 
Whereas in this work, we propose three different types of regularizes to learn robust models using single-step adversarial training methods. The proposed regularizers help to mitigate the effect of gradient masking during single-step adversarial training.
\subsection{Adversarial Sample Generation Methods}
\label{sub_sec:adversarial-sample-generation-methods}
In this subsection we explain methods for generating adversarial samples.  \\ 
\textbf{Fast Gradient Sign Method (FGSM)~\cite{explainingharnessing-iclr-2015}}: Generates adversarial samples based on the first order approximation of the loss function, via performing simple gradient ascent.\\
\begin{equation}
x^* = x + \epsilon. sign\big(\nabla_{x}J(f(x;\theta), y_{true})\big)
\end{equation}
\textbf{Random + Fast Gradient Sign Method (RFGSM)~\cite{ensembleAT-iclr-2018}}: This method adds small random noise before generating an adversarial sample using the FGSM method.
\begin{eqnarray}
x' &=& x + \alpha . sign\big(\mathcal{N}(0^d,I^d)) \\
x^* &=& x' + (\epsilon-\alpha). sign\big(\nabla_{x'}J(f(x';\theta), y_{true})\big)
\end{eqnarray}
Where, $\mathcal{N}$ represents normal distribution\\\\
\textbf{Iterative Fast Gradient Sign Method (IFGSM)~\cite{physicalworld-arxiv-2016}}: In this method, FGSM is applied in an iterative fashion with a small step size ($\alpha$). In our experiments we use $\alpha=\epsilon/steps$. 
\begin{eqnarray}
x^{0} & = & x\\
x^{N+1} & = & x^{N} + \alpha . sign\big(\nabla_{x^{N}}J(f(x^{N};\theta), y_{true})\big)
\end{eqnarray}
\textbf{Projected Gradient Descent (PGD)~\cite{madry-iclr-2018}}: Here the perturbation is initialized with a random point within the allowed $l_\infty$ norm ball and then IFGSM is applied with re-projection.
\begin{eqnarray}
x^{0} & = & x + \mathcal{U}\big(-\epsilon_{step},\epsilon_{step},shape(x)\big) \\
x^{N+1} & = & x^{N} + \epsilon_{step} . sign\big(\nabla_{x^{N}}J(f(x^{N};\theta), y_{true})\big) \\
x^{N+1} & = & clip\big(x^{N+1},min=x-\epsilon,max=x+\epsilon\big) 
\end{eqnarray}
 Where, $\mathcal{U}$ represents uniform distribution.\\\\
\textbf{Projected Gradient Descent with CW loss (PGD-CW)}: Variant of PGD attack which uses C\&W~\cite{robustness-arxiv-2016} loss instead of cross-entropy loss.\\
\\\textbf{Momentum Iterative Fast Gradient Sign Method (MI-FGSM)~\cite{dong2018boosting}}: Introduces momentum into the IFGSM formulation. Here, $\mu$ represents the momentum term and $\alpha$ is set to $\epsilon/steps$. 
\begin{eqnarray}
x^{0} & = & x,~~~p^{0}=0\\
p^{N+1} &=& \mu.p^{N} +  \frac{\nabla_{x^{N}}J(f(x^{N};\theta), y_{true})}{||\nabla_{x^{N}}J(f(x^{N};\theta), y_{true})||_1}\\
x^{N+1} & = & x^{N} + \alpha . sign\big(p^{N+1}\big)
\end{eqnarray}
\\\textbf{DeepFool}: An iterative method proposed by~\cite{deepfool-cvpr-2016}. The method generates an adversarial perturbation based on the linear approximation of the model, that would cause the sample to cross the decision boundary. \\
\\\textbf{Carlini and Wagner (C\&W)}: An iterative method proposed by~\cite{robustness-arxiv-2016}, aims at generating  perturbation with a minimum $l_2$ norm that is sufficient to fool the model i.e., the optimization objective is to find an adversarial perturbation with a minimum $l_2$ norm.   
\subsection{Adversarial training}
\textbf{FGSM Adversarial Training (FGSM-AT)~\cite{atscale-iclr-2017}:}  During FGSM adversarial training, a portion (typically 50\%) of  clean samples in the mini-batch are replaced with their corresponding FGSM adversarial samples. This method is fast and simple, but the resultant models  are not robust to multi-step attacks.
\\\textbf{Ensemble Adversarial Training (EAT)~\cite{ensembleAT-iclr-2018}:} During training, FGSM adversarial samples are generated by the model being trained or by one of the models from a fixed set of normally trained models, chosen at random. Models trained using this method show improvement against adversarial attacks in a black-box setting. Further, models are still susceptible to multi-step attacks in a white-box setting. 
\\\textbf{PGD Adversarial Training (PGD-AT)~\cite{madry-iclr-2018}:}  During training, all the clean samples in the mini-batch are replaced with their corresponding PGD adversarial samples. 
\\\textbf{TRADES~\cite{Zhang2019theoretically}:} During training, an augmented mini-batches containing clean samples and their corresponding adversarial samples are created. These adversarial samples are generated using PGD method with a surrogate loss instead of cross-entropy loss.  Models trained using TRADES and PGD-AT are robust against both single-step and multi-step attacks. However, training time is significantly large when compared to single-step adversarial training methods i.e., FGSM-AT and EAT.
\subsection{Loss surface}\label{subsection:loss_surface}
In this work, we obtained the plot of loss surface~\cite{ensembleAT-iclr-2018} around the data points to illustrate the effect of gradient masking. Loss surface is obtained by varying the input to the model using  Eq.(\ref{equ:loss_surface_eq}).  
\begin{eqnarray}
\label{equ:loss_surface_eq}
x^* &=& x + \epsilon_1 . \delta_1 + \epsilon_2 . \delta_2 
\end{eqnarray}
Where, $\delta_1$ is the sign of the gradient of loss with respect to the input sample, and $\delta_2$ is the sign of the random noise sampled from Normal distribution ($\mathcal{N}$). $\epsilon_1$ and $\epsilon_2$ represent the perturbation size. Loss surface is a 3D plot, where x and y axes represent the perturbation size, and z-axis represents the loss. 
\input{figures/1_main_explaination_plots.tex}
\input{figures/2_main_explaination_decision_surface.tex}
\section{Proposed Approach}\label{sec:proposed_approach}
In this section, we explain the criteria for learning robust models using the adversarial training method~\cite{madry-iclr-2018}. We show that this criterion is not satisfied during the single-step adversarial training. Further, we explain the salient properties that differentiate a robust model from that of a pseudo robust model. Harnessing on these properties, we propose three different types of regularizes that help to learn robust models using single-step adversarial training methods.
\subsection{Criteria for learning robust models}
\label{subsec:criteria_adv}
Madry~\etal\cite{madry-iclr-2018} demonstrated that it is possible to learn robust models using the adversarial training method, if perturbations crafted while training maximize the loss. This objective can be formulated as a mini-max optimization problem (Eq.~\ref{equ:pgd_formaulation}).
\begin{eqnarray}
\label{equ:pgd_formaulation}\displaystyle{\min_{\theta} \Bigg[E_{(x,y)\in D}\bigg[\displaystyle{\max_{\delta\in S}}~J\big(f(x+\delta;\theta),y_{true}\big)\bigg]\Bigg]}
\end{eqnarray}
Where, $D$ is the training dataset, and $S$ is the feasible set $S=\{\delta: ||\delta||_\infty\le\epsilon\}$. At each iteration, we need to find an adversarial perturbation ($||\delta||_\infty\le\epsilon$) that maximizes the model's loss, and further we need to update the model's parameters ($\theta$) so as to minimize the loss on adversarial samples. Madry~\etal\cite{madry-iclr-2018} solves the inner maximization problem by generating adversarial samples using the Projected Gradient Descent method (iterative method). Single-step adversarial training is a special case of mini-max optimization problem (Eq.~\ref{equ:pgd_formaulation}), where the inner maximization is assumed to be achieved by adversarial samples generated by single-step methods. Iterative methods such as PGD, ensure that the generated  perturbations will increase the model's loss, since at each step of the generation process, perturbation with small $\epsilon$ is added to the image. The increase in the loss is not guaranteed when perturbation with high $\epsilon$ is added to the image in a single step. In the next subsection, we show that during the initial stages of single-step adversarial training, the extent of loss maximization achieved by the adversarial samples generated using single-step and multi-step methods are similar. Further, we show that in the later stages of training, single-step adversaries are not able to maximize the loss due to gradient masking effect.   
\subsection{Gradient masking effect}
\label{subsection:effect_of_gradient_masking}
In this subsection, we empirically show that the extent of maximization of loss achieved by FGSM adversaries during the initial stages of single-step adversarial training, is similar to that achieved by PGD (iterative method) adversaries. Further, we show that as training progress, the ability of FGSM samples to maximize the loss diminishes. We train WideResNet-28-10 on CIFAR-10 dataset using  FGSM adversarial training method.
We obtain the plot of cross-entropy loss versus perturbation size ($\epsilon$) of FGSM and PGD attacks, during the initial (at iteration 40 ($\times$100)) and final stages (at iteration 600 ($\times$100)) of training. Bottom-left and bottom-right plots of Fig.~\ref{figure:fgsm_l2_distance} shows the obtained plots. It can be observed that during the initial stage of training, the difference between the average loss on FGSM and PGD adversaries is small (see bottom-left plot of Fig.~\ref{figure:fgsm_l2_distance} for $\epsilon$=8). This implies that the extent of loss maximization achieved by FGSM samples is similar to that achieved by PGD samples. Whereas during the later stage of training, the difference between the average loss on FGSM  and PGD adversaries is large (see bottom-right plot of Fig.~\ref{figure:fgsm_l2_distance} for $\epsilon$=8) i.e., the generated FGSM samples are not able to maximize the training loss. This large difference is due to the gradient masking. During single-step adversarial training, when the model starts to mask the gradient, its decision surface exhibits a sharp curvature near the data points~\cite{ensembleAT-iclr-2018}. This sharp curvature obfuscates the adversarial direction. Single-step adversarial sample generation methods such as FGSM generate adversarial samples based on the linear approximation of the loss function, and gradient masking causes the linear approximation to become unreliable for generating adversarial samples. To illustrate the gradient masking effect, we obtain the loss surface plots. Fig.~\ref{figure:fgsm_decision_surface} shows the loss surface plots obtained during the initial and final stages of FGSM adversarial training. From the left plot of Fig.~\ref{figure:fgsm_decision_surface}, it can be observed that there is no sharp curvature in the loss surface plot of the model obtained during the initial stage of training. Whereas, a sharp curvature can be observed in the loss surface plot of the model obtained during the later stage of training, and this curvature artifact obfuscates the adversarial direction.

 Unlike FGSM adversarial training, during PGD adversarial training, the difference in the average loss on the FGSM and PGD samples is small during the initial and final stages of training (see bottom-left and bottom-right plot of Fig.~\ref{figure:pgd_l2_distance}). Further, from Fig.~\ref{figure:pgd_decision_surface} it can be observed that there is no sharp curvature in the loss surface plots obtained during the initial and final stages of  training. Note that, iterative methods such as PGD generate adversarial samples by adding a small perturbation to the image at every step, and this ensures that the added perturbation will increase the loss. 
\input{figures/3_main_explaination_monotonicity.tex}
\subsection{Salient properties of  robust models}
\label{subsec:salient_properties}
 In this subsection, we bring out the salient properties that differentiate a robust model from that of a pseudo robust model. We train WideResNet-28-10 on CIFAR-10 dataset using FGSM-AT and PGD-AT methods. During training, we obtain the average Euclidean distance between (i) \logits of FGSM and IFGSM adversaries, and (ii) \logits of FGSM and RFGSM adversaries. After training, we obtain the plot of the average cross-entropy loss versus perturbation size of FGSM attack. Following are the salient properties observed in robust models:
 \\\\\textbf{(i) Loss increases monotonically with the increase in perturbation size:}  Column-2 of Fig.~\ref{figure:main_explaination_monotonicity} shows the plot of the average cross-entropy loss versus perturbation size of FGSM attack obtained for the model trained using the PGD-AT method. It can be observed that the average loss increases monotonically with the increase in perturbation size. From column-1 of Fig.~\ref{figure:main_explaination_monotonicity} it can be observed that for the model trained using the FGSM-AT method, the average loss does not increase monotonically with the increase in  perturbation size.     
 \\\textbf{(ii) Similar \logits for adversarial samples generated using different methods:}  The top plot of Fig.~\ref{figure:pgd_l2_distance} shows the average Euclidean distance between \logits of adversarial samples generated using different methods, obtained during PGD-AT. It can be observed that for the entire training duration, the average Euclidean distance is relatively small. Whereas during FGSM-AT, these distances are initially small and become relatively large after a few iterations (Fig.~\ref{figure:fgsm_l2_distance}, top). Note that, during the initial stage of FGSM-AT, the average Euclidean distance between \logits of (i) FGSM and IFGSM samples, and (ii) FGSM and RFGSM samples, are small. Further, these Euclidean distances start to increase when the model starts to mask the gradients (Fig.~\ref{figure:fgsm_l2_distance}, top).
\subsection{Proposed single-step adversarial training with regularization term}
In the previous subsection, we have shown the salient properties that differentiate a robust model from a pseudo robust model. Harnessing on these observations, we propose three different types of regularizers which help to learn robust models using single-step adversarial training methods. The proposed regularizers penalize the model for masking gradients, and this enables the inclusion of useful single-step adversarial samples during the entire training process. 
\subsubsection{Single-step Adversarial Training with Regularizer-1 (SAT-R1)}\label{subsubsection:sat_r1}
In subsection~\ref{subsec:salient_properties}, we observed that during single-step adversarial training, the average Euclidean distance between the \logits of FGSM and IFGSM adversaries increase drastically. Whereas during PGD-AT, this distance is relatively small and does not increase as training proceeds. Based on these observations, we include a penalty term in the training loss (Eq.~\ref{equ:proposed_penalty_r1}) to minimize the distance between \logits of FGSM and IFGSM adversaries of clean samples during single-step adversarial training. 
\begin{equation}
\begin{split}
\mathcal{L}oss=& \frac{1}{m}\sum_{i=1}^{m}J(f(x^i_{fgsm};\theta),y^i_{true}) \\&+ \lambda\frac{1}{k}\sum_{j=1}^{k}\big\|{g(x^j_{fgsm}) - g(x^j_{ifgsm})}\big\|_{2}^{2} \end{split}
\label{equ:proposed_penalty_r1}
\end{equation}
In Eq. (\ref{equ:proposed_penalty_r1}), the first term corresponds to the classification task, and  the second term represents the proposed regularization. Further, $\lambda$ represents the regularization weighting factor and $k$ represents the number of adversaries generated using IFGSM. During training, when the model starts to mask the gradients,  the proposed penalty term causes training loss to increase (since the distance between \logits of FGSM and IFGSM adversarial pair increases). This behavior of the proposed penalty term helps in mitigating the effect of gradient masking, and thus enables the generation of stronger adversaries while training. In  section~\ref{sec:experiments} we show that $k$=1 (i.e., penalty is imposed on one FGSM and IFGSM adversarial pair of a clean sample in the mini-batch) is sufficient to learn robust models. This means that only \textit{one} adversarial sample in the mini-batch is generated using an iterative method and the remaining adversarial samples are generated using non-iterative method. Further, we show that adversarial training with mini-batches containing $one$ IFGSM and $m$ FGSM samples without the proposed regularizer, does not improve the model's robustness significantly. The result of this ablation experiment is shown in section~\ref{sec:experiments}.
\subsubsection{Single-step Adversarial Training with Regularizer-2 (SAT-R2)}
\label{subsubsection:sat_r2}
In section~\ref{subsec:salient_properties}, we showed that when the model starts to mask gradients, then the Euclidean distance between \logits of FGSM and RFGSM adversaries of a clean sample becomes large. Based on this observation, we introduce a regularization term in the training loss (Eq.\ref{equ:proposed_penalty_r2}) that penalizes the effect of gradient masking during single-step adversarial training.
\begin{equation}
\begin{split}
\mathcal{L}oss=& \frac{1}{m}\sum_{i=1}^{m}J(f(x^i_{fgsm};\theta),y^i_{true}) \\&+ \lambda\frac{1}{m}\sum_{j=1}^{m}\big\|{g(x^j_{fgsm}) - g(x^j_{rfgsm})}\big\|_{2}^{2} 
\end{split}
\label{equ:proposed_penalty_r2}
\end{equation}
In Eq.(\ref{equ:proposed_penalty_r2}), the first term corresponds to the classification loss, and the second term represents the proposed regularization.  During training, if the model starts to mask gradients, then the Euclidean distance between \logits of FGSM and RFGSM adversaries of clean samples increases, and this in turn causes the training loss (Eq.~\ref{equ:proposed_penalty_r2}) to increase. This behavior of the proposed regularizer prevents the model from masking gradients. Note that, adversarial training with RFGSM or with both RFGSM and FGSM samples does not improve the model's robustness significantly. The results of these experiments are shown in section~\ref{sec:experiments}.
\input{tables/mnist_architecture_table.tex}
\input{tables/mnist_eat_table.tex}
\subsubsection{Single-step Adversarial Training with Regularizer-3 (SAT-R3)}
\label{subsubsection:sat_r3}
In subsection~\ref{subsec:salient_properties}, we demonstrated that for a robust model, loss on the FGSM adversarial samples increases monotonically with the increase in  perturbation size, and this behavior is not observed in the model trained using the single-step adversarial training method. Based on this observation, we propose a regularisation term which enforces the model's loss to increase  monotonically with the increase in  perturbation size. Eq.~(\ref{equ:proposed_penalty_r3}) represents the training loss.
\begin{equation}
\mathcal{L}oss = loss_{\epsilon_{High}} +  \lambda.max\big(0, loss_{\epsilon_{Low}} - \tau.{loss_{\epsilon_{High}}}\big)  
\label{equ:proposed_penalty_r3}
\end{equation}
 Where, $loss_{\epsilon_{Low}}$ and $loss_{\epsilon_{High}}$ represent the average loss on FGSM adversarial samples with perturbation size of $\epsilon_{Low}$ and $\epsilon_{High}$ respectively. The first term corresponds to the classification task, and the second term represents the monotonic loss constraint. During training, we ensure $loss_{\epsilon_{Low}} < loss_{\epsilon_{High}}$ by enforcing ($loss_{\epsilon_{Low}}/loss_{\epsilon_{High}}) < \tau$, where $\tau<1$. For stability purpose, we consider ($loss_{\epsilon_{Low}} - \tau.loss_{\epsilon_{High}}) < 0$. During training, if the model starts to mask gradients, the second term in  Eq.~(\ref{equ:proposed_penalty_r3}) becomes greater than zero, and this causes training loss to increase. This behavior of the proposed loss constraint, explicitly prevents the model from generating weaker adversaries during adversarial training. Also, note that the monotonicity  of loss is enforced for the allowed perturbation range [0, $\epsilon_{High}$] i.e., the maximum value of $\epsilon_{High}$ is restricted based on perceptual constraints.  
\input{tables/mnist_L_infty_table.tex}
\input{tables/cifar10_L_infty_table.tex}
\input{tables/imagenet_L_infty_table.tex}

\section{Experiments}\label{sec:experiments}
In this section, we show the performance of models trained using the proposed training methods against adversarial attacks in white-box and black-box settings. We perform sanity tests described in~\cite{obfuscated-gradients,carlini2019evaluating}  to ensure that models trained using the proposed regularizers are robust, and do not exhibit obfuscated gradients. Since, models exhibiting gradient masking or obfuscated gradients are not robust against adversarial attacks~\cite{obfuscated-gradients}. Code for the proposed approach is available at \url{https://github.com/val-iisc/SAT-Rx}.\\
\\\textbf{Dataset}: We show results on MNIST~\cite{lecun1998mnist}, CIFAR-10~\cite{cifar_10_dataset} and ImageNet-subset (100 classes)~\cite{imagenet-ijcv-2015} datasets. For ImageNet-subset, we randomly choose 100 classes. We use LeNet+ (refer to table~\ref{table:eat_mnsit_architecure}), WideResNet-28-10 (WRN-28-10)~\cite{BMVC2016_87} and ResNet-18~\cite{resnet-2015} for MNIST, CIFAR-10 and ImageNet-subset datasets respectively. Images are pre-processed to be in [0,1] range. For data-augmentation, horizontal flip and random crop are performed for CIFAR-10 and ImageNet-subset datasets.
\\\textbf{Training methods:} We compare the proposed training methods with Normal training (NT), FGSM Adversarial Training (FGSM-AT)~\cite{atscale-iclr-2017}, Ensemble Adversarial Training (EAT)~\cite{ensembleAT-iclr-2018}, PGD Adversarial Training (PGD-AT)~\cite{madry-iclr-2018}, and TRADES~\cite{Zhang2019theoretically}. Refer table~\ref{table:ensemble_setup} for details on the experimental setup used for EAT.  
\\\textbf{Attacks}: We show the performance of models trained using different training methods against $l_\infty$ and $l_2$ attacks. For $l_\infty$ norm-bounded attacks, we use FGSM~\cite{explainingharnessing-iclr-2015}, IFGSM~\cite{physicalworld-arxiv-2016}, MI-FGSM~\cite{dong2018boosting}, PGD~\cite{madry-iclr-2018} and PGD-CW attacks. For $l_2$ attacks, we use DeepFool~\cite{deepfool-cvpr-2016} and C\&W~\cite{robustness-arxiv-2016} attacks. We follow Madry~\etal\cite{madry-iclr-2018} for attack parameters. For $l_\infty$ attacks, we limit $\epsilon$ to 0.3, 8/255 and 8/255  for MNIST, CIFAR-10, and ImageNet-subset datasets respectively. Note that, DeepFool and C\&W attacks measure the robustness of the model based on the $l_2$ norm of the generated adversarial perturbation.
\\\textbf{Hyper-parameters}: For SAT-R1, we set ($\lambda$, $k$) to (0.2, 1), (0.2, 1) and (0.05, 1) for MNIST, CIFAR-10 and ImageNet-subset datasets respectively. For SAT-R2, we set $\lambda$ to 5, 25 and 3 for MNIST, CIFAR-10 and ImageNet-subset datasets respectively. For SAT-R3, we set ($\lambda$, $\tau$) to (1, 0.4), (1, 0.6) and (1, 0.6) for MNIST, CIFAR-10 and ImageNet-subset datasets respectively. 
\subsection{Performance against $l_\infty$ attacks}\label{subsec:l_infty_attacks}
We train models on MNIST, CIFAR-10, and ImageNet-subset datasets using the proposed single-step adversarial training methods SAT-R1, SAT-R2, and SAT-R3. Further, we also train models using NT, FGSM-AT, EAT, PGD-AT, and TRADES methods. Models are trained for 20, 100 and 100 epochs on MNIST, CIFAR-10, and ImageNet-subset datasets respectively. Table~\ref{table:mnist_linfty_performance},~\ref{table:cifar10_linfty_performance} and~\ref{table:imgenet_linfty_performance} shows the performance of these models against single-step and multi-step attacks in white-box and black-box settings. For black-box attacks, a normally trained model is used for generating adversarial samples, and these generated adversarial samples are tested on the target model. Typically, the model used for generating adversarial samples is referred to as a ``source model" or ``substitute model".
\\\textbf{White-box setting:} From tables~\ref{table:mnist_linfty_performance},~\ref{table:cifar10_linfty_performance} and~\ref{table:imgenet_linfty_performance}, it can be observed that models trained using single-step adversarial training methods (i.e., FGSM-AT, EAT) are susceptible to multi-step attacks. Whereas models trained using PGD-AT, TRADES, SAT-R1, SAT-R2, and SAT-R3 are robust against both single-step and multi-step attacks. Note that, PGD-AT and TRADES use iterative methods for the generation of adversarial samples, due to which training time is significantly high. Unlike PGD-AT and TRADES, the proposed methods SAT-R1, SAT-R2, and SAT-R3 use non-iterative method for crafting adversarial samples. Further, we obtain the plot of test-set recognition accuracy of models trained using the proposed regularizers for PGD attack with increasing steps. Fig.~\ref{figure:acc_vs_pgdsteps} shows the obtained plot, it can observed that the model's accuracy saturates with increase in steps of PGD attack. We obtain this plot to verify that the model's performance does not degrade significantly with the increase in the number of iteration/steps of adversarial attack~\cite{engstrom2018evaluating}. 
\\\textbf{Black-box setting:} The last four columns of table~\ref{table:mnist_linfty_performance}, ~\ref{table:cifar10_linfty_performance} and~\ref{table:imgenet_linfty_performance} shows the performance of models in a black-box setting. It can be observed that the performance of models trained using  PGD-AT, TRADES, SAT-R1, SAT-R2, and SAT-R3, in a black-box setting is better than that in a white-box setting. Note that, the model trained on CIFAR-10 dataset using FGSM-AT is more susceptible to adversarial attack in a black-box setting than in a white-box setting.    
\subsection{Performance against $l_2$ attacks}\label{subsec:l_2_attacks}
DeepFool and C\&W attacks belongs to a class of attacks that generate adversarial perturbations without norm constraints. These attacks aim to generate adversarial perturbations with a minimum $l_2$ norm, that is just sufficient to fool the classifier. The average $l_2$ norm of the generated perturbations indicates the robustness of the model. For an undefended classifier, perturbations with a small $l_2$ norm is sufficient to fool the classifier. Whereas, for a robust classifier, perturbations with relatively large $l_2$ norm, are required to fool the classifier. Table~\ref{table:deepfool_cw} shows the performance of models against DeepFool and C\&W attacks. Fooling Rate (FR) represents the percentage of test-set samples that are misclassified. It can be observed that for models trained using PGD-AT, SAT-R1, SAT-R2, and SAT-R3, the average $l_2$ norm of the generated perturbations is relatively high.
\subsection{Ablation study}\label{subsec:ablation_study}
We perform ablation study to show the significance of the proposed regularizers. For the ablation study, we train LeNet+ on MNIST dataset.
\subsubsection{Ablation study on SAT-R1}\label{subsubsec:ablation_sat_r1}
\textbf{Ablation-R1-1:} SAT-R1 with $\lambda$=0. We train LeNet+ on MNIST dataset using SAT-R1 with $\lambda$=0, and cross-entropy loss imposed on both IFGSM and FGSM samples. We perform this experiment to show that the gain in the robustness of the model is due to the proposed regularizer, and not due to the inclusion of one IFGSM sample. Table~\ref{table:ablations} shows the performance of the model trained using this method, and it can be observed that the model is not robust to multi-step attacks.
\\\textbf{Ablation-R1-2:} SAT-R1 with mini-batch containing $k$  IFGSM  and $m$ (mini-batch size) FGSM samples. We train LeNet+ using SAT-R1 with different values of $k$, to show that it is sufficient to impose the proposed regularizer on a single pair of FGSM and IFGSM sample in a mini-batch. Column-1 of Fig.~\ref{figure:ablation_r1} shows the plot of accuracy of the model on the PGD validation set, trained using SAT-R1 with different values of $k$. It can be observed that $k$=1 is sufficient to learn robust models using SAT-R1.
\\\textbf{Ablation-R1-3:} Adversarial training with mini-batches containing $k$  IFGSM and $m$ FGSM samples without the proposed regularizer. In this experiment, cross-entropy loss is imposed on FGSM and IFGSM adversarial samples. We train LeNet+ on MNIST dataset and during training, we generate $k$ (expressed in terms of original mini-batch size $m$)  IFGSM and $m$ FGSM samples. We perform this experiment to show that without the proposed regularizer, at least 40\% of the samples should be generated using IFGSM method, so as to learn robust models. Column-2 of Fig.~\ref{figure:ablation_r1} shows the plot of accuracy of the model on the PGD validation set, for different values of $k$.
It can be observed that for $k$= 40\%, there is a significant improvement in the model's robustness. This implies that at least 40\% of the samples should be generated using the IFGSM method so as to learn robust models without the proposed regularizer.
\input{tables/deepfool_cw.tex}
\input{tables/ablation.tex}
\input{figures/8_ablation_r1.tex}
\subsubsection{Ablation study on SAT-R2}\label{subsubsec:ablation_sat_r2}
\textbf{Ablation-R2-1:} SAT-R2 with $\lambda$=0. We train LeNet+ on MNIST dataset using SAT-R2 with $\lambda$=0, and cross-entropy loss is imposed on both FGSM and RFGSM samples. We perform this experiment to show that the gain in the robustness of models trained using SAT-R2 is not due to the inclusion of RFGSM samples. 
\\\textbf{Ablation-R2-2:} Adversarial training with RFGSM samples. We train LeNet+ on MNIST dataset using the adversarial training method, and during training, adversarial samples are generated using RFGSM. We perform this experiment to show that adversarial training with RFGSM samples does not improve the model's robustness significantly.  

\subsubsection{Ablation study on SAT-R3}\label{subsubsec:ablation_sat_r3}
\textbf{Ablation-R3-1:} SAT-R3 with $\lambda$=0. We train LeNet+ on MNIST dataset using SAR-R3 with $\lambda$=0. We perform this experiment to show that adversarial training with only FGSM samples does not improve the model's robustness.  
\\\textbf{Ablation-R3-2:} SAT-R3 with $\lambda$=0 and cross-entropy is imposed on FGSM samples with the perturbation size of $\epsilon_{Low}$ and $\epsilon_{High}$. We train LeNet+ on MNIST dataset using SAT-R3 with $\lambda$=0, and cross-entropy imposed on FGSM samples with the perturbation size of $\epsilon_{Low}$ and $\epsilon_{High}$. We perform this experiment to show that adversarial training with mini-batches containing FGSM samples with the perturbation size of $\epsilon_{Low}$ and $\epsilon_{High}$, does not improve the model's robustness. 

Table~\ref{table:ablations}  shows the performance of models trained using the above training methods, and it can be observed that there is no significant improvement in the model's robustness against multi-step attacks.
\subsection{Sanity tests}\label{subsec:obfuscated_gradeint_detection}
We perform sanity test described in Carlini~\etal\cite{carlini2019evaluating} to verify the robustness of models trained using the proposed method, and to rule out obfuscated gradients. Athalye~\etal\cite{obfuscated-gradients}  showed that certain defense methods unintentionally or intentionally cause models to exhibit obfuscated gradients. Further, a method to break such defense methods was proposed. This implies that models exhibiting obfuscated or masked gradients are not robust. We perform the following sanity tests to verify the robustness of models trained using the proposed regularizers:
\\\textbf{(i) Verify multi-step attacks perform better than single-step attacks}: For robust models, iterative attacks should be stronger than non-iterative attacks in a white-box setting. Table~\ref{table:mnist_linfty_performance},~\ref{table:cifar10_linfty_performance} and ~\ref{table:imgenet_linfty_performance}, shows the performance of models against single-step and multi-step attacks, and  it can be observed that iterative attacks (IFGSM and PGD) are stronger than non-iterative attack (FGSM) for models trained using SAT-R1, SAT-R2, and SAT-R3 methods. 
\\\textbf{(ii) Verify white-box attacks perform better than black-box attacks}: For any model, white-box attacks should be stronger than black-box attacks. In black-box setting, partial or no knowledge of the deployed model is available to the attacker, and hence black-box attacks should be weaker than white-box attacks.  From table~\ref{table:mnist_linfty_performance},~\ref{table:cifar10_linfty_performance} and ~\ref{table:imgenet_linfty_performance}, it can be observed that white-box attacks are stronger than black-box attacks on models trained using  SAT-R1, SAT-R2, and SAT-R3 methods. 
\\\textbf{(iii) Verify for large perturbation size, model's accuracy reach levels of random guessing}: Typically, the model's performance degrades drastically for attacks with a large perturbation size ($\epsilon$). We obtain the plot of recognition accuracy (\%) of the model on PGD test set for different values of attack perturbation size. Fig.~\ref{figure:proposed_acc_vs_eps} shows the plot obtained for models trained using the proposed regularizers, it can be  observed that the recognition accuracy (\%) of the model is zero for PGD attack with large perturbation size.
\\\textbf{(iv) Verify increase in the perturbation size strictly increases attack success rate}: From Fig.~\ref{figure:proposed_acc_vs_eps}, it can be observed that the PGD attack success rate (success rate = 100 - accuracy) increases with the increase in the distortion bound i.e., perturbation size of PGD attack.
\input{figures/5_acc_vs_steps.tex}
\subsection{Loss trend}\label{subsec:loss_vs_eps_plots}
In this subsection, we obtain the plot of average loss on the test set versus perturbation size of FGSM and PGD attacks. Fig.~\ref{figure:proposed_loss_vs_fgsm_pgd_eps} shows the plots obtained for models trained using the proposed regularizers. It can be observed that the average loss increases monotonically with the increase in perturbation size of FGSM and PGD attacks. Further, it can be observed that the difference between the loss on FGSM and PGD samples is relatively small, even for higher perturbation size ($\epsilon$). Whereas, this difference would be large for models exhibiting masked gradients.
\subsection{Loss surface}\label{subsec:decision_surface_plots}
In this subsection, we obtain the loss surface plots for models trained using the proposed regularizers. Fig.~\ref{figure:proposed_loss_vs_fgsm_pgd_eps} shows the obtained plots. It can be observed that there is no sharp curvature near the decision points. Whereas, sharp curvature near data points can be observed for models exhibiting masked gradients (see column-2 plot of Fig.~\ref{figure:fgsm_decision_surface}).
\input{figures/4_accuracy_versus_eps.tex}

\input{tables/complexity.tex}
\input{tables/sequential_training.tex}
\vspace{-0.5cm}
\subsection{Complexity}\label{subsec:complexity}
In this subsection, we discuss the complexity of different adversarial training methods. The adversarial sample generation  is the major bottleneck for adversarial training methods. Typically, adversarial sample generation involves computation of gradient of the loss with respect to the input image. This requires one or more forward and backward propagation through the network. We define a metric $FBP$, which corresponds to one Forward and Backward propagation through the network. Further, we use $FBP$ to express the complexity of the adversarial sample generation process. Table~\ref{table:training_complexity} summaries the complexity involved in generating adversarial samples during different adversarial training methods. For FGSM-AT, which uses single-step method for generating adversaries, requires only one forward and backward propagation through the network to generate adversaries. For EAT, a fixed set of pre-trained source models are required along with the model being trained for generating FGSM adversarial samples. Therefore, additional training of source models is involved. PGD-AT and TRADES methods, which use iterative methods for generating adversaries, require multiple forward and backward propagation through the network, and this causes a significant increase in the training time. The proposed training methods use single-step methods for generating adversaries, and this requires only one forward and backward propagation through the network, except for SAT-R1 where one sample in the mini-batch is generated using an iterative method. Among the proposed methods, complexity of SAT-R1$>$SAT-R2$>$SAT-R3. SAT-R1 requires one IFGSM and `m' FGSM samples. The generation of one adversarial sample using iterative method is the bottleneck for SAT-R1. SAT-R2 requires generation of `m' FGSM and `m' R-FGSM samples. Whereas, SAT-R3 requires `m' FGSM samples with perturbation size of $\epsilon_{Low}$ and  `m' FGSM samples with perturbation size of $\epsilon_{High}$. We use computational trick to reduce the time required for generating these adversarial samples i.e., sign of the gradient of loss required for generating adversarial samples with perturbation size of $\epsilon_{High}$ and $\epsilon_{Low}$ is same. Therefore computational complexity of SAT-R3 is lesser than SAT-R2. 
\input{figures/6_loss_vs_fgsm_pgd_eps.tex}
\input{figures/7_proposed_decision_surface.tex}
\vspace{-0.3cm}
\subsection{Comparison of SAT-R1, SAT-R2 and SAT-R3}\label{subsec:comparison}
 In terms of computational complexity, SAT-R1$>$SAT-R2$>$SAT-R3. The performance of models trained using SAT-R1 and SAT-R2 are similar. Whereas the performance of models trained using SAT-R3 is relatively less superior than compared to models trained using SAT-R1 and SAT-R2. Among the proposed regularizers, SAT-R1 and SAT-R2 are suitable when performance is of importance, and SAT-R3 is suitable for cases where training time is the bottleneck e.g., training on large datasets. 
 
 Further, we demonstrate that it is possible to boost the performance of models trained using SAT-R3, by either pre-training or fine-tuning these models  using SAT-R1 or SAT-R2 for few epochs. Table~\ref{table:sequential_training} shows the performance of LeNet+ trained on MNIST dataset using this approach. It can be observed that there is an improvement in the model's robustness against multi-step attacks (PGD-40 and PGD-100) when compared to the model trained using SAT-R3 only. 
\section{Discussion and conclusion}\label{sec:discussion_conclusion}
 Adversarial training, a straightforward solution to defend  model against adversarial attacks, shows promising results. However, models trained using the existing single-step adversarial training converge to degenerative minima where the model appears to be (pseudo) robust. Though multi-step adversarial training methods such as PGD adversarial training and TRADES methods help to learn adversarially robust models, they are computationally expensive. 

In this work, we have proposed three different types of regularizers for single-step adversarial training. The proposed regularizers harness the salient properties of robust models to mitigate gradient masking effect, and help to learn robust models using single-step methods in a computationally efficient manner. Unlike models trained using the existing single-step adversarial training methods, models trained using the proposed methods are robust against both single-step and multi-step attacks.

\bibliographystyle{IEEEtran}
\bibliography{IEEEabrv,main.bib}
%
\begin{IEEEbiography}[{\includegraphics[width=0.8in,height=1.25in,clip,keepaspectratio]{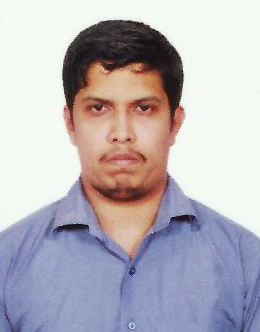}}]{B.S. Vivek} is a research student at Video Analytics Lab, CDS, Indian Institute of Science, Bangalore. He received his B.E. from M.S. Ramaiah Institute of Technology, Bangalore. His research interest includes computer vision and machine learning.
\end{IEEEbiography}
\begin{IEEEbiography}[{\includegraphics[width=0.8in,height=1.25in,clip,keepaspectratio]{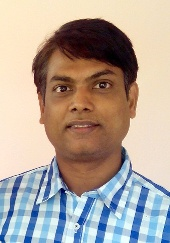}}]{\\ \\ R. Venkatesh Babu} received his Ph.D from Dept. of Electrical Engineering, IISc, Bangalore. Thereafter, he held postdoctoral positions at NTNU, Norway and IRISA/INRIA, France. Subsequently, he worked as a research fellow at NTU, Singapore. He is currently an Associate Professor at Dept. of CDS and convener of VAL, IISc. His interests span vision, image/video processing, ML, and multimedia.
\end{IEEEbiography}



\end{document}

%% file: figures/1_main_explaination_plots.tex
\begin{figure*}[h!]
    \begin{minipage}[t]{0.49\textwidth}
        \centering
        {\includegraphics[width=\textwidth]{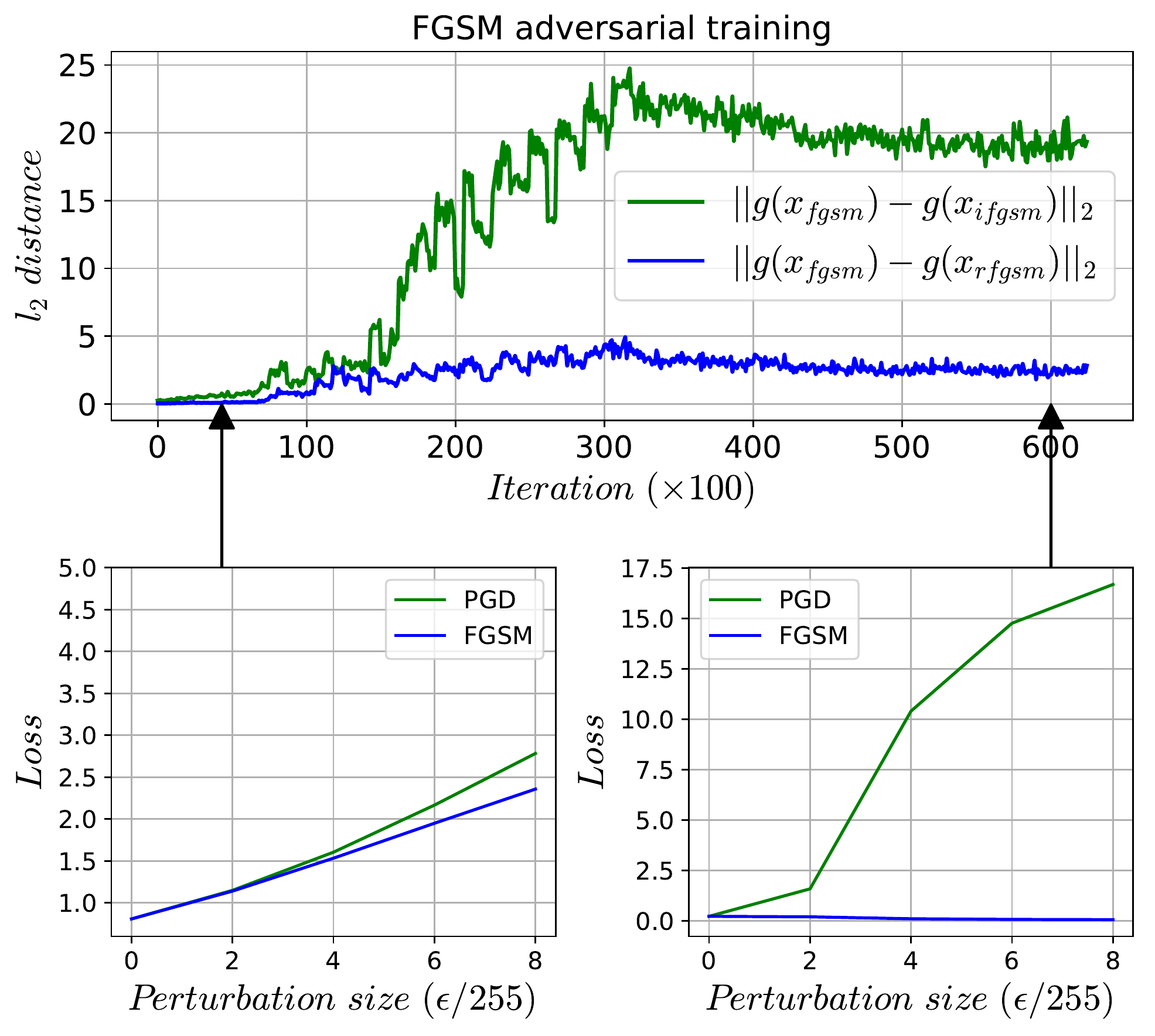}}
        \caption{\textbf{Top}: Plot of average $l_2$ distance between  
(i) \logits of FGSM and IFGSM adversaries, and (ii) \logits of FGSM and RFGSM adversaries of clean samples, obtained for the model trained on CIFAR-10 dataset using FGSM adversarial training method. Observe the increases in the $l_2$ distance after $\sim$80 iteration ($\times$100). \textbf{Bottom}: Plot of the average loss of the model versus perturbation size of PGD and FGSM attacks. Bottom-left: Plot obtained at iteration 40 ($\times$100), Bottom-right: Plot obtained at iteration 600 ($\times$100). Observe the gradient masking effect in the bottom-right plot i.e., difference between the average loss on PGD and FGSM samples is large for $\epsilon$=8/255.}
        \label{figure:fgsm_l2_distance}
    \end{minipage}
    \hspace{0.2cm}%
    \begin{minipage}[t]{0.49\textwidth}
        \centering
        {\includegraphics[width=\textwidth]{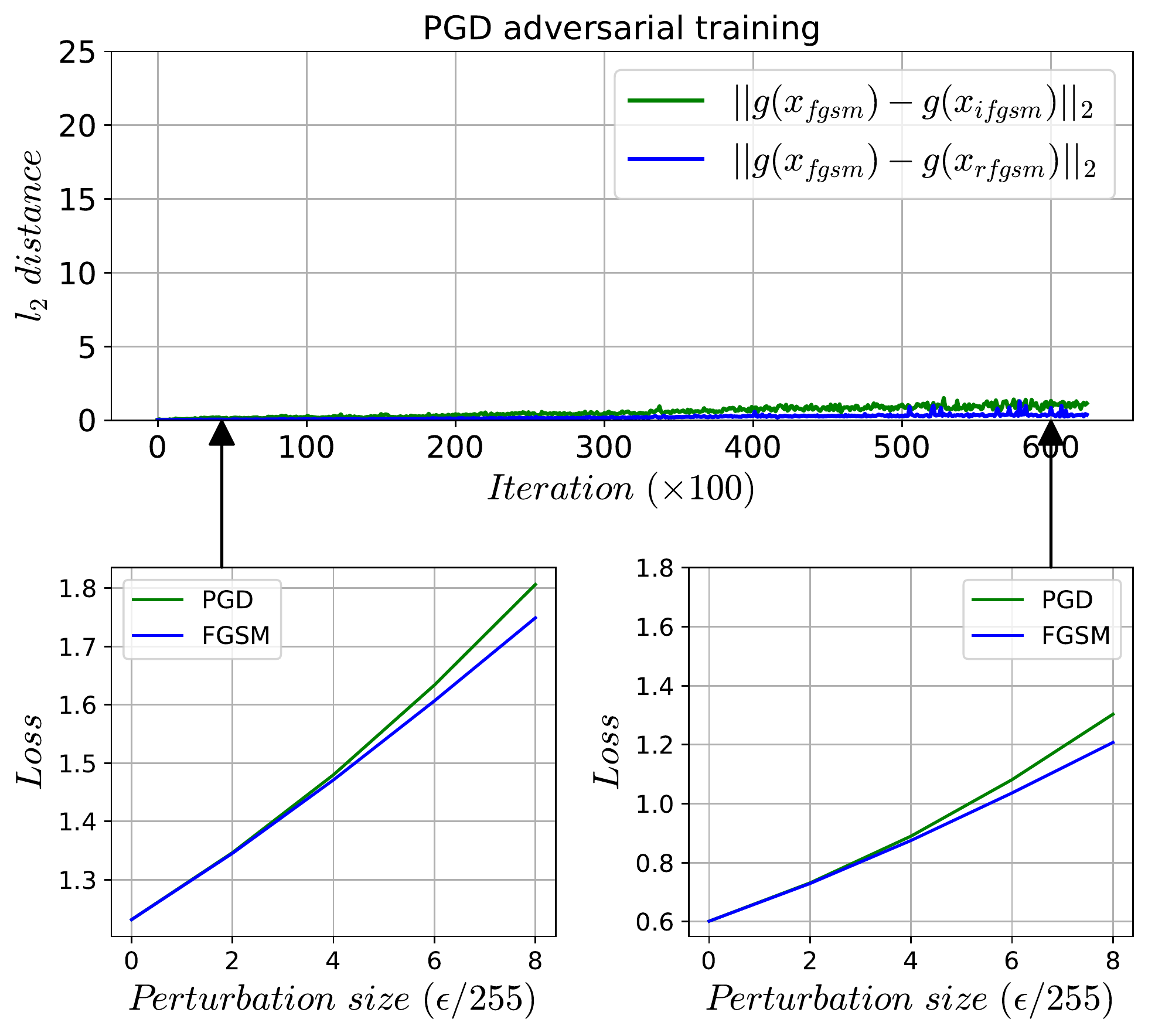}}
        \caption{\textbf{Top}: Plot of average $l_2$ distance between (i) \logits of FGSM and IFGSM adversaries, and (ii) \logits of FGSM and RFGSM adversaries of clean samples, obtained for the model trained on CIFAR-10 dataset using PGD adversarial training method. Observe that for the entire training duration, average $l_2$ distance is relatively small. \textbf{Bottom}: Plot of the average loss of the model versus perturbation size of PGD and FGSM attacks. Bottom-left: Plot obtained at iteration 40 ($\times$100). Bottom-right: Plot obtained at iteration 600 ($\times$100).}
        \label{figure:pgd_l2_distance}
    \end{minipage}
\end{figure*}

%% file: figures/2_main_explaination_decision_surface.tex
\begin{figure*}[h!]
    \begin{minipage}[t]{0.49\textwidth}
        \centering
        {\includegraphics[width=\textwidth]{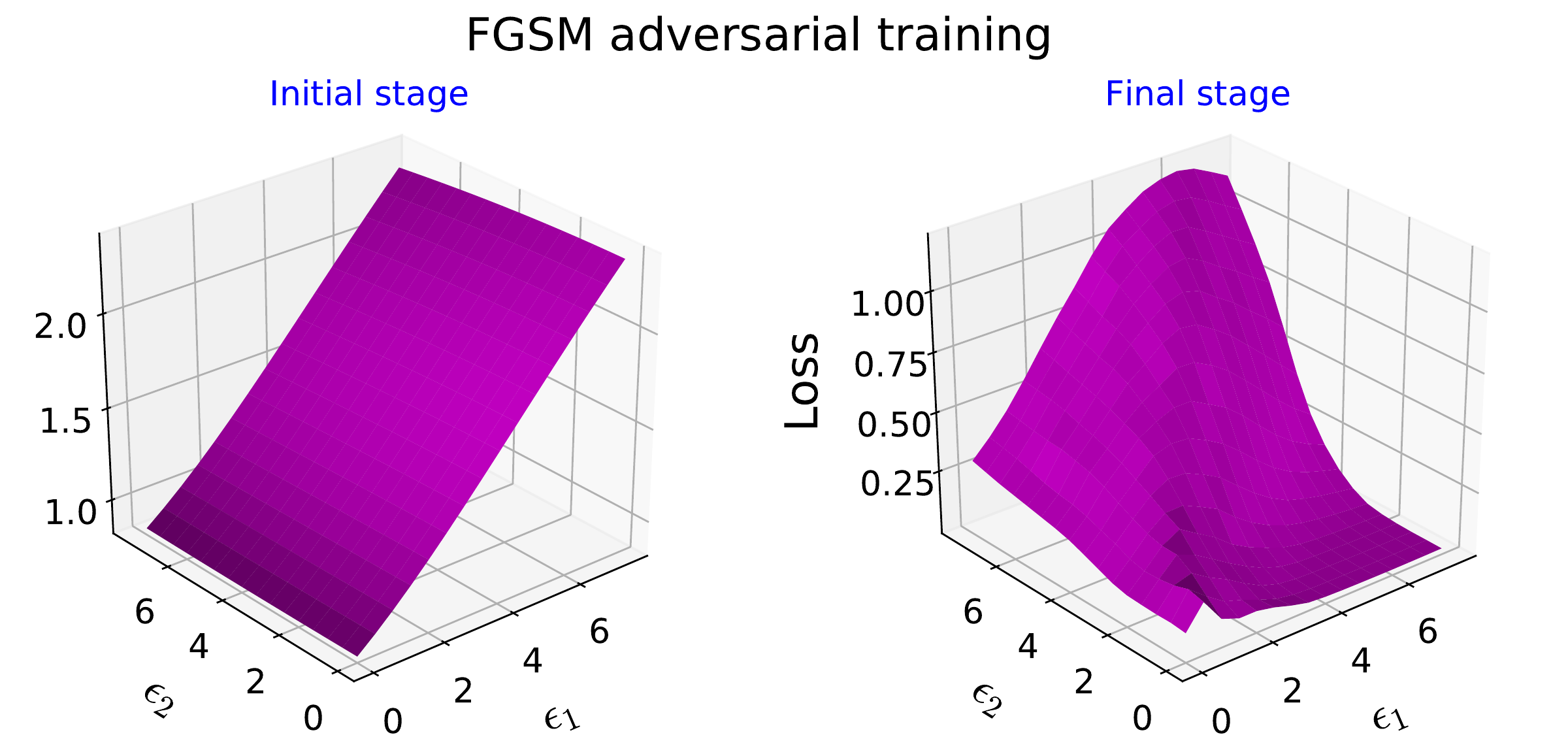}}
        \caption{Loss surface plot of the model trained using FGSM adversarial training method. \textbf{Left}: loss surface obtained during the initial stage of training. \textbf{Right}: loss surface obtained during the final stage of training. Please refer to section~\ref{subsection:loss_surface} for details on loss surface plot.}
        \label{figure:fgsm_decision_surface}
    \end{minipage}
    \hspace{0.2cm}%
    \begin{minipage}[t]{0.49\textwidth}
        \centering
        {\includegraphics[width=\textwidth]{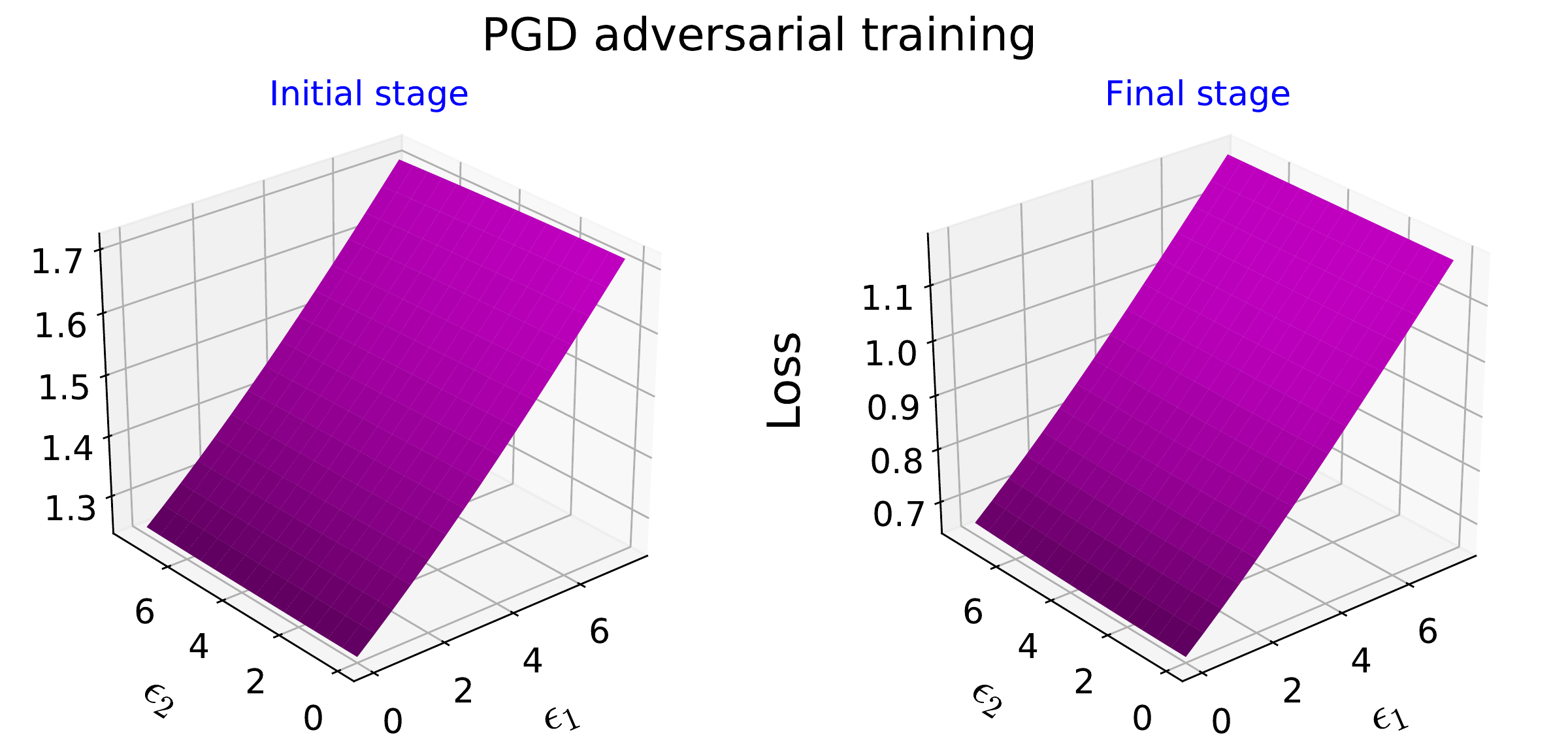}}
        \caption{Loss surface plot of the model trained using PGD adversarial training method. \textbf{Left}: loss surface obtained during the initial stage of training. \textbf{Right}: loss surface obtained during the final stage of training. Please refer to section~\ref{subsection:loss_surface} for details on loss surface plot.}
        \label{figure:pgd_decision_surface}
    \end{minipage}
\end{figure*}

%% file: figures/3_main_explaination_monotonicity.tex
\begin{figure}[t!]
        \centering
        {\includegraphics[width=\linewidth]{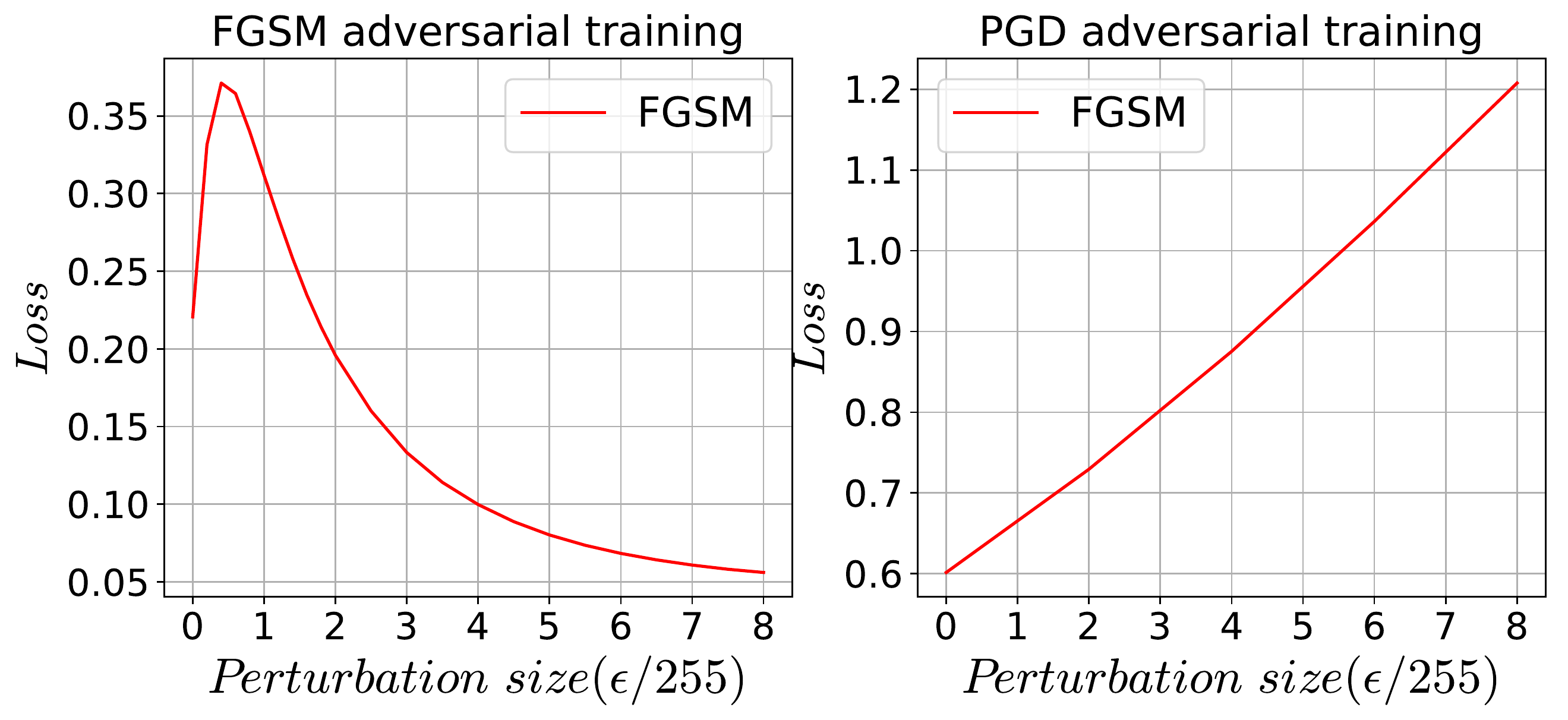}}
        \vspace{-0.5cm}
        \caption{Plot of loss versus perturbation size of FGSM attack, obtained for the model trained using single-step and multi-step adversarial training methods. \textbf{Left}: FGSM adversarially trained model. \textbf{Right}: PGD adversarially trained model. Observe that, for PGD adversarially trained model loss increases monotonically with the increase in perturbation size.}
        \label{figure:main_explaination_monotonicity}
        \vspace{-0.3cm}
\end{figure}

%% file: tables/mnist_architecture_table.tex
\begin{table*}[]
    \centering
    \caption{Architecture of networks used for Ensemble Adversarial Training (EAT) on MNIST dataset.}
    \label{table:eat_mnsit_architecure}
    \begin{tabular}{c|c|c|c|cc} \toprule
    \textbf{LeNet+} & \textbf{A} & \textbf{B} & \textbf{C} & \textbf{D} \\ \midrule
    Conv(32,5,5) + Relu & Conv(64,5,5) + Relu  & Dropout(0.2)         & Conv(128,3,3) + Tanh   & \multirow{2}{*}{$\Big\{$} FC(300) +Relu \multirow{2}{*}{$\Big\}\times4$}\\
    MaxPool(2,2) & Conv(64,5,5) + Relu  & Conv(64,8,8)  + Relu & MaxPool(2,2)           & Dropout(0.5)  \\
    Conv(64,5,5) + Relu & Dropout(0.25)        & Conv(128,6,6) + Relu & Conv(64,3,3) + Tanh    & FC + Softmax  \\
    MaxPool(2,2) & FC(128) + Relu       & Conv(128,5,5) + Relu & MaxPool(2,2)           &   \\ 
    FC(1024) + Relu & Dropout(0.5)        & Dropout(0.5)        & FC(128) + Relu         &   \\
    FC + Softmax    & FC + Softmax         & FC + Softmax         & FC + Softmax           &   \\ \bottomrule
    \end{tabular}
    \vspace{-0.3cm}
\end{table*}

%% file: tables/mnist_eat_table.tex
\begin{table}[t]
\caption{Setup used for Ensemble Adversarial Training (EAT). Please refer to  table~\ref{table:eat_mnsit_architecure} for details on  models used for MNSIT dataset.}
\label{table:ensemble_setup}
\centering
\setlength\tabcolsep{3pt}
\resizebox{1.01\linewidth}{!}{
\begin{tabular}{c|c|c}\toprule
                                 & \textbf{Network to be trained}      & \textbf{Pre-trained Models }    \\ \toprule
                                 & LeNet+ (ens-A)                  & LeNet+, A               \\  
MNIST                            & LeNet+ (ens-B)                  & LeNet+, B               \\  
                                 & LeNet+ (ens-C)                  & A, B                    \\ \midrule
                                 & WRN-28-10 (ens-A)               & WRN-28-10, ResNet-34            \\ 
CIFAR-10                         & WRN-28-10 (ens-B)               & WRN-28-10, VGG-19               \\ 
                                 & WRN-28-10 (ens-C)               & ResNet-34, VGG-19              \\ \midrule
                                 & ResNet-18 (ens-A)               & ResNet-18, ResNet-34            \\ 
ImageNet Subset                  & ResNet-18 (ens-B)               & ResNet-18, VGG-11               \\ 
                                 & ResNet-18 (ens-C)               & ResNet-34, VGG-11              \\ \bottomrule                                 
\end{tabular}
}
\end{table}

%% file: tables/mnist_L_infty_table.tex
\begin{table*}[]
\caption{MNIST: Recognition accuracy (\%) of models trained on MNIST dataset using different training methods in white-box and black-box settings. Rows represent  training methods and columns represent  attack methods. For all the attacks, $\epsilon$ is set to 0.3. For PGD and PGD-CW attacks, $\epsilon_{step}$ is set to 0.01.  The number of iterations/steps for multi-step attacks is set to 100. In black-box setting,  model C and D are used for generating adversarial samples.}
\vspace{-0.3cm}
\label{table:mnist_linfty_performance}
\centering
\begin{tabular}{@{}lrrrrrr|rr|rr@{}}
\toprule
\multicolumn{6}{c}{\textbf{White-box}} & \multicolumn{1}{l}{\textbf{}} & \multicolumn{4}{c}{\textbf{Black-box}} \\ \midrule
\multicolumn{1}{c}{\textbf{Training}} & \multicolumn{1}{l}{\textbf{}} & \multicolumn{4}{c}{\textbf{Attacks}} & \multicolumn{1}{l}{\textbf{}} & \multicolumn{2}{c}{\textbf{C}} & \multicolumn{2}{c}{\textbf{D}} \\ \cmidrule(lr){3-6} \cmidrule(l){8-11} 
\multicolumn{1}{c}{\textbf{method}} & \multicolumn{1}{l}{\textbf{Clean}} & \multicolumn{1}{l}{\textbf{FGSM}} & \multicolumn{1}{l}{\textbf{IFGSM}} & \multicolumn{1}{l}{\textbf{PGD}} & \multicolumn{1}{l}{\textbf{PGD-CW}} & \multicolumn{1}{l}{\textbf{}} & \multicolumn{1}{l}{\textbf{FGSM}} & \multicolumn{1}{l}{\textbf{MI-FGSM}} & \multicolumn{1}{l}{\textbf{FGSM}} & \multicolumn{1}{l}{\textbf{MI-FGSM}} \\ \cmidrule(r){1-6} \cmidrule(l){8-11}
NT & 99.24 & 11.50 & 0.24 & 0.00 & 0.00 &  & 22.28 & 9.57 & 51.31 & 44.65 \\
FGSM-AT & 99.29 & 85.86 & 17.02 & 3.95 & 5.32 &  & 91.81 & 88.20 & 90.12 & 88.92 \\
EAT ens-A & 99.37 & 79.63 & 5.39 & 0.34 & 0.41 &  & 89.33 & 83.44 & 92.14 & 90.78 \\
EAT ens-B & 99.31 & 84.12 & 2.36 & 0.06 & 0.08 &  & 86.79 & 77.79 & 91.62 & 88.57 \\
EAT ens-C & 99.43 & 80.08 & 2.87 & 0.05 & 0.11 &  & 85.82 & 76.44 & 92.21 & 89.52 \\
PGD-AT & 98.41 & 95.56 & 92.53 & 91.18 & 91.34 &  & 95.74 & 95.52 & 95.68 & 95.51 \\
TRADES & 98.70 & 96.25 & 94.96 & 93.60 & 93.69 &  & 96.43 & 95.93 & 96.32 & 96.05 \\\cmidrule(r){1-6} \cmidrule(l){8-11} 
SAT-R1 & 98.15 & 93.42 & 91.45 & 88.76 & 88.63 &  & 93.99 & 93.81 & 94.04 & 94.23 \\
 & $\pm$0.05 & $\pm$0.07 & $\pm$0.26 & $\pm$0.26 & $\pm$0.27 &  & $\pm$0.09 & $\pm$0.07 & $\pm$0.06 & $\pm$0.07 \\\cmidrule(r){1-6} \cmidrule(l){8-11}
SAT-R2 & 98.23 & 94.36 & 92.17 & 90.03 & 89.73 &  & 94.64 & 94.50 & 94.72 & 94.91 \\
 &$\pm$0.12 & $\pm$0.09 & $\pm$0.40 & $\pm$0.32 & $\pm$0.29 &  & $\pm$0.06 & $\pm$0.04 & $\pm$0.04 & $\pm$0.05 \\\cmidrule(r){1-6} \cmidrule(l){8-11}
SAT-R3 & 98.79 & 94.44 & 88.01 & 82.88 & 83.50 &  & 94.52 & 94.48 & 95.03 & 95.11 \\
 & $\pm$0.24 & $\pm$0.36 & $\pm$0.39 & $\pm$0.60 & $\pm$0.43 &  & $\pm$0.25 & $\pm$0.30 & $\pm$0.22 & $\pm$0.24 \\ \bottomrule
\end{tabular}
\end{table*}

%% file: tables/cifar10_L_infty_table.tex
\begin{table*}[]
\caption{CIFAR-10: Recognition accuracy (\%) of models trained on CIFAR-10 dataset using different training methods in white-box and black-box settings. Rows represent  training methods and columns represent  attack methods. For all the attacks, $\epsilon$ is set to 8/255.  For PGD and PGD-CW attacks, $\epsilon_{step}$ is set to 2/255.  The number of iterations/steps for multi-step attacks is set to 20. In black-box setting, VGG-11 and DenseNet-BC-100 are used for generating adversarial samples.}
\vspace{-0.3cm}
\label{table:cifar10_linfty_performance}
\centering
\begin{tabular}{@{}lrrrrrr|rr|rr@{}}
\toprule
\multicolumn{6}{c}{\textbf{White-box}} & \multicolumn{1}{l}{\textbf{}} & \multicolumn{4}{c}{\textbf{Black-box}} \\ \midrule
\multicolumn{1}{c}{\textbf{Training}} & \multicolumn{1}{l}{\textbf{}} & \multicolumn{4}{c}{\textbf{Attacks}} & \multicolumn{1}{l}{\textbf{}} & \multicolumn{2}{c}{\textbf{VGG-11}} & \multicolumn{2}{c}{\textbf{DenseNet-BC-100}} \\ \cmidrule(lr){3-6} \cmidrule(l){8-11} 
\multicolumn{1}{c}{\textbf{method}} & \multicolumn{1}{l}{\textbf{Clean}} & \multicolumn{1}{l}{\textbf{FGSM}} & \multicolumn{1}{l}{\textbf{IFGSM}} & \multicolumn{1}{l}{\textbf{PGD}} & \multicolumn{1}{l}{\textbf{PGD-CW}} & \multicolumn{1}{l}{\textbf{}} & \multicolumn{1}{l}{\textbf{FGSM}} & \multicolumn{1}{l}{\textbf{MI-FGSM}} & \multicolumn{1}{l}{\textbf{FGSM}} & \multicolumn{1}{l}{\textbf{MI-FGSM}} \\ \cmidrule(r){1-6} \cmidrule(l){8-11} 
NT & 94.75 & 28.16 & 0.02 & 0.00 & 0.00 &  & 48.46 & 31.61 & 39.58 & 28.50 \\
FGSM-AT & 94.04 & 98.54 & 0.04 & 0.01 & 0.02 &  & 78.70 & 76.35 & 86.90 & 86.42 \\
EAT ens-A & 92.92 & 59.56 & 19.05 & 7.11 & 8.02 &  & 80.99 & 80.35 & 87.52 & 87.72 \\
EAT ens-B & 92.75 & 63.40 & 6.26 & 0.85 & 0.88 &  & 83.72 & 83.60 & 89.01 & 89.26 \\
EAT ens-C & 93.11 & 59.74 & 15.36 & 3.73 & 3.96 &  & 84.21 & 84.35 & 88.95 & 89.35 \\
PGD-AT & 86.30 & 53.96 & 51.30 & 47.92 & 48.01 &  & 82.67 & 82.82 & 84.57 & 84.58 \\
TRADES & 86.92 & 56.34 & 53.00 & 49.92 & 50.13 &  & 82.83 & 82.97 & 84.77 & 84.87 \\\cmidrule(r){1-6} \cmidrule(l){8-11} 
SAT-R1 & 84.68 & 56.85 & 50.36 & 46.81 & 47.36 &  & 80.76 & 80.86 & 82.77 & 83.06 \\
 & $\pm$1.40 & $\pm$0.08 & $\pm$0.13 & $\pm$0.66 & $\pm$0.22 &  & $\pm$1.70 & $\pm$1.67 & $\pm$1.67 & $\pm$1.71 \\\cmidrule(r){1-6} \cmidrule(l){8-11} 
SAT-R2 & 83.51 & 56.38 & 52.90 & 49.07 & 49.22 &  & 80.05 & 80.13 & 82.21 & 82.26 \\
 & $\pm$0.14 & $\pm$0.07 & $\pm$0.49 & $\pm$1.14 & $\pm$0.66 &  & $\pm$0.26 & $\pm$0.26 & $\pm$0.48 & $\pm$0.20 \\\cmidrule(r){1-6} \cmidrule(l){8-11} 
SAT-R3 & 80.95 & 50.63 & 43.96 & 40.13 & 40.30 &  & 76.47 & 76.53 & 78.99 & 79.34 \\
 & $\pm$0.75 & $\pm$2.24 & $\pm$2.08 & $\pm$2.82 & $\pm$2.40 &  & $\pm$1.49 & $\pm$1.47 & $\pm$1.91 & $\pm$1.94 \\ \bottomrule
\end{tabular}
\vspace{-0.2cm}
\end{table*}

%% file: tables/imagenet_L_infty_table.tex
\begin{table*}[]
\caption{ImageNet-subset: Recognition accuracy (\%) of models trained on ImageNet-subset dataset using different training methods in white-box and black-box settings. Rows represent  training methods and columns represent  attack methods. For all the attacks, $\epsilon$ is set to 8/255. For PGD and PGD-CW attacks, $\epsilon_{step}$ is set to 2/255.  The number of iterations/steps for multi-step attacks is set to 20. In black-box setting, VGG-13 and ResNet-50 are used for generating adversarial samples.}
\vspace{-0.3cm}
\label{table:imgenet_linfty_performance}
\centering
\begin{tabular}{@{}lrrrrrr|rr|rr@{}}
\toprule
\multicolumn{6}{c}{\textbf{White-box}} & \multicolumn{1}{l}{\textbf{}} & \multicolumn{4}{c}{\textbf{Black-box}} \\ \midrule
\multicolumn{1}{c}{\textbf{Training}} & \multicolumn{1}{l}{\textbf{}} & \multicolumn{4}{c}{\textbf{Attacks}} & \multicolumn{1}{l}{\textbf{}} & \multicolumn{2}{c}{\textbf{VGG-13}} & \multicolumn{2}{c}{\textbf{ResNet-50}} \\ \cmidrule(lr){3-6} \cmidrule(l){8-11} 
\multicolumn{1}{c}{\textbf{method}} & \multicolumn{1}{l}{\textbf{Clean}} & \multicolumn{1}{l}{\textbf{FGSM}} & \multicolumn{1}{l}{\textbf{IFGSM}} & \multicolumn{1}{l}{\textbf{PGD}} & \multicolumn{1}{l}{\textbf{PGD-CW}} & \multicolumn{1}{l}{\textbf{}} & \multicolumn{1}{l}{\textbf{FGSM}} & \multicolumn{1}{l}{\textbf{MI-FGSM}} & \multicolumn{1}{l}{\textbf{FGSM}} & \multicolumn{1}{l}{\textbf{MI-FGSM}} \\ \cmidrule(r){1-6} \cmidrule(l){8-11} 
NT & 78.74 & 4.68 & 0.04 & 0.00 & 0.00 &  & 47.08 & 26.64 & 71.28 & 67.30 \\
FGSM-AT & 77.10 & 40.51 & 12.46 & 2.84 & 4.08 &  & 72.48 & 72.90 & 75.78 & 74.66 \\
EAT ens-A & 76.04 & 39.21 & 8.00 & 2.92 & 3.20 &  & 69.64 & 69.74 & 73.78 & 72.92 \\
EAT ens-B & 76.44 & 40.83 & 9.00 & 3.56 & 3.64 &  & 71.20 & 71.46 & 74.72 & 73.70 \\
EAT ens-C & 76.66 & 41.72 & 8.92 & 3.32 & 3.26 &  & 70.70 & 71.20 & 74.66 & 73.70 \\
PGD-AT & 68.18 & 40.50 & 36.48 & 33.72 & 33.62 &  & 66.84 & 66.96 & 67.66 & 67.44 \\
TRADES & 67.32 & 42.10 & 36.80 & 34.34 & 34.53 &  & 65.93 & 66.51 & 67.30 & 67.10 \\\cmidrule(r){1-6} \cmidrule(l){8-11} 
SAT-R1 & 61.60 & 38.20 & 32.67 & 30.52 & 29.37 &  & 60.42 & 60.47 & 60.89 & 60.84 \\
 & $\pm$0.36 & $\pm$0.17 & $\pm$0.14 & $\pm$0.34 & $\pm$0.70 &  & $\pm$0.40 & $\pm$0.34 & $\pm$0.37 & $\pm$0.45 \\\cmidrule(r){1-6} \cmidrule(l){8-11} 
SAT-R2 & 61.56 & 37.79 & 32.83 & 30.35 & 28.89 &  & 60.53 & 60.53 & 61.09 & 60.93 \\
 & $\pm$1.61 & $\pm$0.77 & $\pm$0.51 & $\pm$0.67 & $\pm$0.97 &  & $\pm$1.47 & $\pm$1.51 & $\pm$1.62 & $\pm$1.57 \\\cmidrule(r){1-6} \cmidrule(l){8-11} 
SAT-R3 & 67.87 & 38.73 & 30.51 & 27.67 & 28.13 &  & 65.86 & 67.09 & 67.37 & 66.44 \\
 & $\pm$0.75 & $\pm$1.60 & $\pm$1.32 & $\pm$1.43 & $\pm$1.49 &  & $\pm$1.35 & $\pm$1.78 & $\pm$1.30 & $\pm$1.15 \\ \bottomrule
\end{tabular}
\end{table*}

%% file: tables/deepfool_cw.tex
\begin{table*}[]
\caption{Performance of models trained  using different training methods against DeepFool and CW attacks. These attack methods measure the robustness of the model based on the average $l_2$ norm of the generated perturbations, higher the better. FR defines the percentage of test set samples that has been misclassified. Note that,  models trained using PGD-AT, SAT-R1, SAT-R2 and SAT-R3, requires perturbations with relatively large $l_2$ norm to fool the classifier.}
\centering
\label{table:deepfool_cw}
\begin{tabular}{l|rr|rr|rr|rr|rr|rr}
\toprule
\multirow{3}{*}{\textbf{Method}} & \multicolumn{4}{c|}{\textbf{MNIST}}                              & \multicolumn{4}{c|}{\textbf{CIFAR-10}}                            & \multicolumn{4}{c}{\textbf{ImageNet-subset}}                           \\  \cmidrule(l){2-13} 
                              & \multicolumn{2}{c|}{\textbf{DeepFool}} & \multicolumn{2}{c|}{\textbf{CW}} & \multicolumn{2}{c|}{\textbf{DeepFool}} & \multicolumn{2}{c|}{\textbf{CW}} & \multicolumn{2}{c|}{\textbf{DeepFool}} & \multicolumn{2}{c}{CW} \\ \cmidrule(l){2-13} 
                              & \textbf{FR}       & \textbf{Mean} $l_2$       & \textbf{FR}    & \textbf{Mean} $l_2$    & \textbf{FR}       & \textbf{Mean} $l_2$       & \textbf{FR}    & \textbf{Mean} $l_2$    & \textbf{FR}       & \textbf{Mean} $l_2$       & \textbf{FR}    & \textbf{Mean} $l_2$    \\ \midrule
NT & 99 & 1.83 & 100 & 1.68 & 96 & 0.20 & 100 & 0.12 &94 & 0.39 & 100 & 0.32 \\
FGSM-AT & 99 & 2.81 & 100 & 1.96 & 96 & 0.25 & 100 & 0.10 &92 & 0.70 & 100 & 0.23 \\ 
EAT ens-A & 99 & 2.69 & 100 & 1.88 & 95 & 0.69 & 100 & 0.61 &91	& 1.23 & 100 & 0.89 \\ 
EAT ens-B & 99 & 2.68 & 100 & 1.86 & 95 & 0.63 & 100 & 0.70 &91	& 1.34 & 100 & 1.03 \\ 
EAT ens-C & 100 & 2.60 & 100 & 1.84 & 95 & 0.77 & 100 & 0.70&92	& 1.37 & 100 & 1.09 \\ 
PGD-AT & 86 & 4.62 & 100 & 3.69 & 92 & 1.22 & 100 & 0.88 &90	& 2.04 & 100 & 1.94 \\ \midrule
SAT-R1 & 89 & 4.51 & 100 & 3.34 & 92 & 1.11 & 100 & 0.84 &91	& 1.61 & 100 & 1.74 \\ 
SAT-R2 & 89 & 5.28 & 94 & 3.55 & 90 & 1.58 & 100 & 1.01 &89	& 1.52 & 98 & 1.36 \\ 
SAT-R3 & 95 & 3.01 & 100 & 2.13 & 90 & 0.98 & 100 & 0.87 &89	& 1.95 & 100 & 1.64 \\ \bottomrule
\end{tabular}
\vspace{-0.4cm}
\end{table*}

%% file: tables/ablation.tex
\begin{table}[]
\centering
\caption{\textbf{Ablation study:} Recognition accuracy (\%) of models trained on MNIST dataset, for adversarial attacks in white-box setting. Please refer to section~\ref{subsec:ablation_study} for details on training methods.}
\label{table:ablations}
\begin{tabular}{lllll}
\toprule
 &  & \multicolumn{3}{c}{\textbf{White-box}} \\ \midrule
\multicolumn{1}{c}{\textbf{Training}} &  & \multicolumn{3}{c}{\textbf{Adversarial attacks}} \\ \cmidrule(l){3-5} 
\multicolumn{1}{c}{\textbf{Method}} & \textbf{Clean} & \textbf{FGSM} & \textbf{PGD} & \textbf{PGD} \\ 
                &                               &                               & \textbf{steps=40}                      & \textbf{steps=100}\\\midrule
Ablation-R1-1   & \multicolumn{1}{r}{97.71}     & \multicolumn{1}{r}{91.35}     & \multicolumn{1}{r}{16.48}     &\multicolumn{1}{r}{0.97}     \\
SAT-R1          & \multicolumn{1}{r}{98.15}     & \multicolumn{1}{r}{93.42}     & \multicolumn{1}{r}{90.26}     &\multicolumn{1}{r}{88.76}      \\
                & \multicolumn{1}{r}{$\pm$0.05} & \multicolumn{1}{r}{$\pm$0.07} & \multicolumn{1}{r}{$\pm$0.16} &\multicolumn{1}{r}{$\pm$0.26}      \\ \midrule
Ablation-R2-1   & \multicolumn{1}{r}{95.39}     & \multicolumn{1}{r}{61.60}     & \multicolumn{1}{r}{34.15}     &\multicolumn{1}{r}{33.41}      \\
Ablation-R2-2   & \multicolumn{1}{r}{93.92}     & \multicolumn{1}{r}{57.06}     & \multicolumn{1}{r}{29.31}     &\multicolumn{1}{r}{28.76}      \\
SAT-R2          & \multicolumn{1}{r}{98.23}     & \multicolumn{1}{r}{94.36}     & \multicolumn{1}{r}{91.22}     &\multicolumn{1}{r}{90.03}      \\
                & \multicolumn{1}{r}{$\pm$0.12} & \multicolumn{1}{r}{$\pm$0.09} & \multicolumn{1}{r}{$\pm$0.37} &\multicolumn{1}{r}{$\pm$0.32}      \\ \midrule
Ablation-R3-1   & \multicolumn{1}{r}{97.62}     & \multicolumn{1}{r}{92.52}     & \multicolumn{1}{r}{1.89}      &\multicolumn{1}{r}{0.78}      \\
Ablation-R3-2   & \multicolumn{1}{r}{97.56}     & \multicolumn{1}{r}{72.35}     & \multicolumn{1}{r}{1.53}      &\multicolumn{1}{r}{0.63}      \\
SAT-R3          & \multicolumn{1}{r}{98.79}     & \multicolumn{1}{r}{94.44}     & \multicolumn{1}{r}{86.48}     &\multicolumn{1}{r}{82.88}      \\
                & \multicolumn{1}{r}{$\pm$0.24} & \multicolumn{1}{r}{$\pm$0.36} & \multicolumn{1}{r}{$\pm$0.42} &\multicolumn{1}{r}{$\pm$0.60}      \\ \midrule
\end{tabular}
\vspace{-0.3cm}
\end{table}

%% file: figures/8_ablation_r1.tex
\begin{figure}[t!]
        \centering
        {\includegraphics[width=\linewidth]{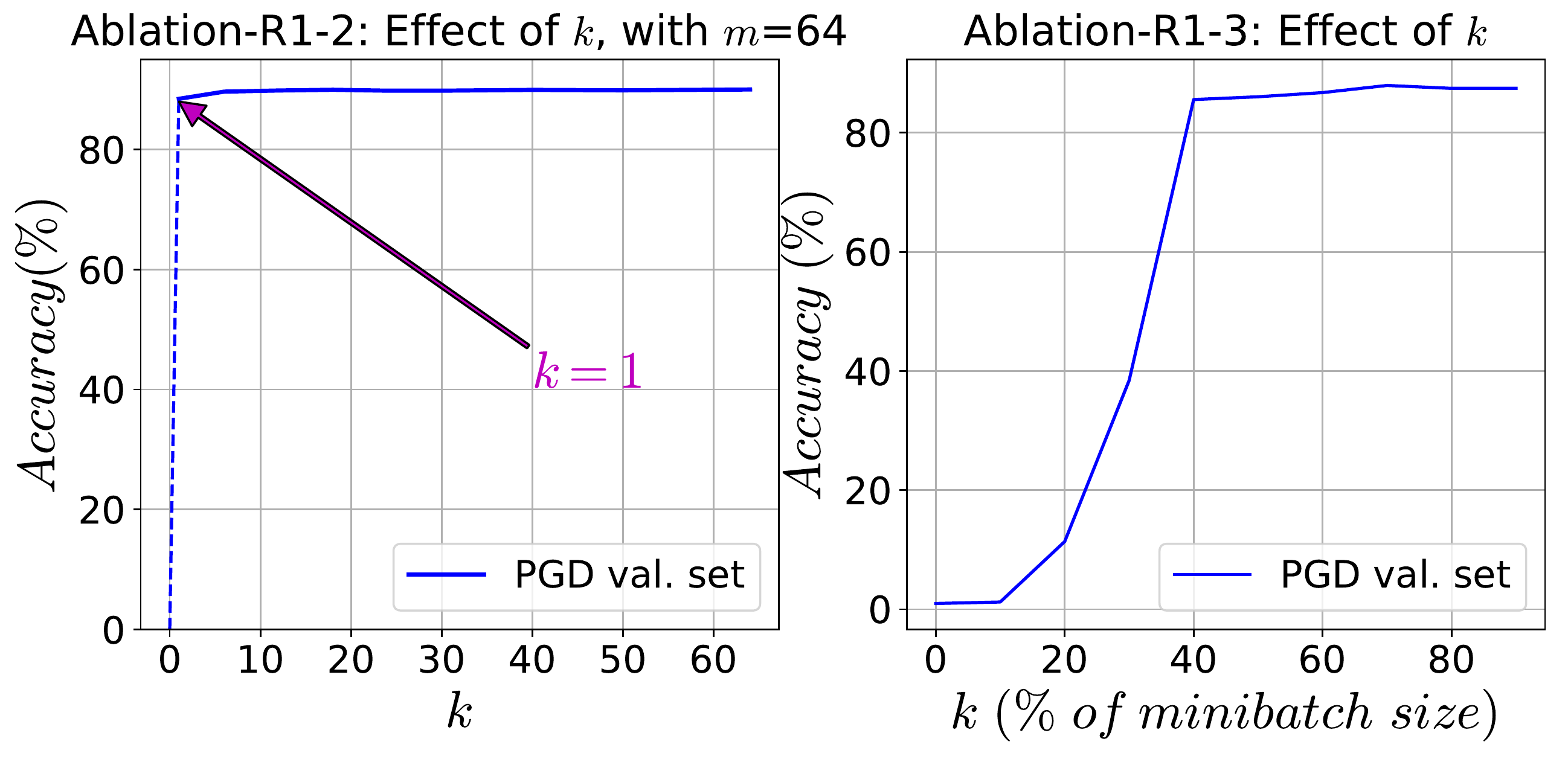}}
        \vspace{-0.5cm}
        \caption{Column-1: plot of accuracy of the model trained using SAT-R1 with different values of $k$. Column-2: plot of accuracy of the model trained using Ablation-R1-3 (without regularizer) with different values of $k$ (expressed in term of percentage (\%) of mini-batch size)}
        \label{figure:ablation_r1}
        \vspace{-0.3cm}
\end{figure}

%% file: figures/5_acc_vs_steps.tex
\begin{figure}[t!]
\centering    
	\includegraphics[width=\linewidth]{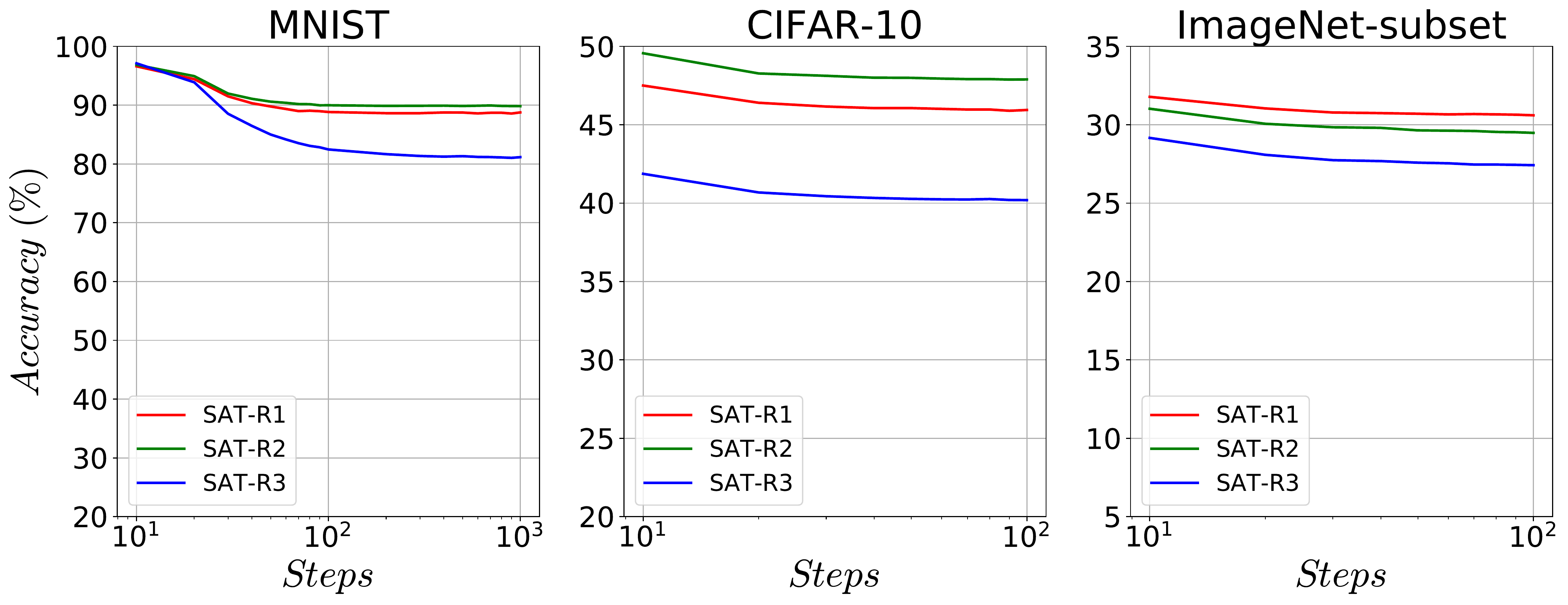}
	\vspace{-0.4cm}
	\caption{Plot of recognition accuracy (\%) versus steps/iteration of PGD attack with fixed perturbation size ($\epsilon$), obtained for models trained using SAT-R1, SAT-R2 and SAT-R3. We set  $\epsilon$ to 0.3, 8/255 and 8/255 for MNIST, CIFAR-10 and ImageNet-subset datasets respectively. Note that, $x$-axis is in logarithmic scale. Observe the saturation of model's accuracy for PGD attack with large steps/iteration.} 
	\label{figure:acc_vs_pgdsteps}   
\end{figure}

%% file: figures/4_accuracy_versus_eps.tex
\begin{figure}[t!]
        \centering
        {\includegraphics[width=\linewidth]{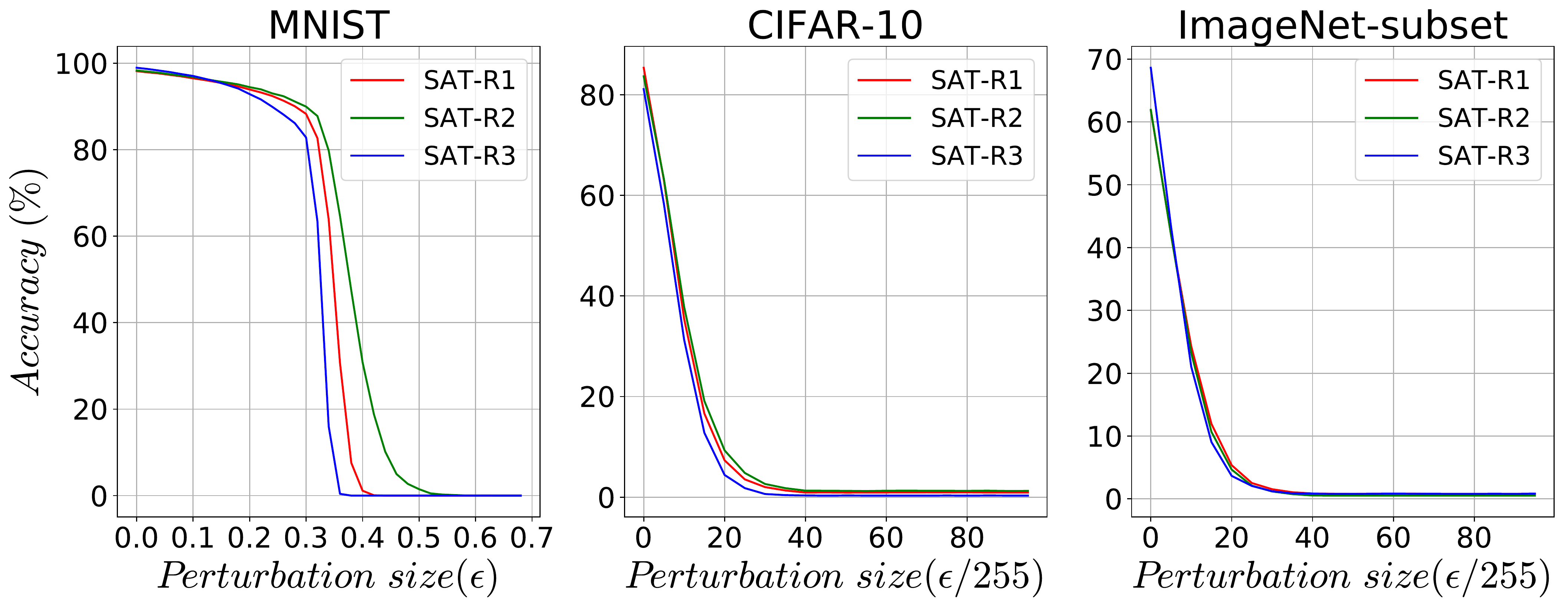}}
        \vspace{-0.4cm}
        \caption{Plot of the recognition accuracy (\%) of the model versus perturbation size ($\epsilon$) of  PGD attack, obtained for models trained using the proposed regularizers.  Note that, accuracy of the model is zero for larger perturbation size.}
        \label{figure:proposed_acc_vs_eps}
\end{figure}

%% file: tables/complexity.tex
\begin{table}[h]
\caption{Complexity of adversarial training methods. $FBP$ is equal to the number of iterations/steps used for generating adversarial samples, and this is dataset dependent. * For SAT-1, one sample in the mini-batch is generated using IFGSM, and the corresponding $FBP$ value is mentioned inside the parenthesis.}
\label{table:training_complexity}
\setlength\tabcolsep{3pt}
\resizebox{0.98\linewidth}{!}{
\begin{tabular}{@{}lcccc@{}}
\toprule
\multicolumn{1}{c}{\textbf{Training}} & \multicolumn{3}{c}{\textbf{FBP}}                                                   & \textbf{Pre-trained} \\ \cmidrule(lr){2-4}
\multicolumn{1}{c}{\textbf{method}}   & \multicolumn{1}{l}{\textbf{MNSIT}} & \textbf{CIFAR-10} & \multicolumn{1}{l}{\textbf{ImagNet-Subset}}    & \textbf{models}      \\ \midrule
FGSM-AT                      & 1                         & 1        & 1                                                 & No          \\
EAT                          & 1                         & 1        & 1                                                 & Yes         \\
PGD-AT                       & 40                        & 7        & 7                                                 & No          \\
TRADES                       & 40                        & 7        & 7                                                 & No          \\\midrule
SAT-R1*                      & 1 (40)                    & 1 (7)    & 1 (7)                                             & No          \\
SAT-R2                       & 1                         & 1        & 1                                                 & No          \\
SAT-R3                       & 1                         & 1        & 1                                                 & No          \\ \bottomrule
\end{tabular}
}
\end{table}

%% file: tables/sequential_training.tex
\begin{table}[h!]
\centering
\caption{\textbf{Boosting performance of SAT-R3}: Recognition accuracy (\%) of LeNet+ trained on MNIST dataset using the proposed regularizers in sequential manner. \textbf{SAT-R1 + SAT-R3:} Model is pre-trained using SAT-R1 for the first 5 epochs, and is trained using SAT-R3 for the next 15 epochs. \textbf{SAT-R3 + SAT-R2:} Model is trained using SAT-R3 for 15 epochs, and is fine-tuned using SAT-R2 for 5 epochs.}
\label{table:sequential_training}
\begin{tabular}{lllll}
\toprule
 &  & \multicolumn{3}{c}{\textbf{White-box}} \\ \midrule
\multicolumn{1}{c}{\textbf{Training}} &  & \multicolumn{3}{c}{\textbf{Adversarial attacks}} \\ \cmidrule(l){3-5} 
\multicolumn{1}{c}{\textbf{Method}} & \textbf{Clean} & \textbf{FGSM} & \textbf{PGD} & \textbf{PGD} \\ 
                &                               &                               & \textbf{steps=40}                      & \textbf{steps=100}\\\midrule
SAT-R3          & \multicolumn{1}{r}{98.79}     & \multicolumn{1}{r}{94.44}     & \multicolumn{1}{r}{86.48}     &\multicolumn{1}{r}{82.88}      \\
 (20 epochs)      & \multicolumn{1}{r}{$\pm$0.24} & \multicolumn{1}{r}{$\pm$0.36} & \multicolumn{1}{r}{$\pm$0.42} &\multicolumn{1}{r}{$\pm$0.60}      \\ \midrule
\multicolumn{5}{l}{\textbf{Sequential training}} \\\midrule               
SAT-R1 + SAT-R3 & \multicolumn{1}{r}{98.80} & \multicolumn{1}{r}{94.42} & \multicolumn{1}{r}{89.84} &\multicolumn{1}{r}{87.52}      \\
(5+15 epochs)   & \multicolumn{1}{r}{$\pm$0.09} & \multicolumn{1}{r}{$\pm$0.16} & \multicolumn{1}{r}{$\pm$0.70} &\multicolumn{1}{r}{$\pm$1.21}      \\ \midrule      
SAT-R3 + SAT-R2 & \multicolumn{1}{r}{98.10} & \multicolumn{1}{r}{93.83} & \multicolumn{1}{r}{89.08} &\multicolumn{1}{r}{87.44}      \\
(15+5 epochs)   & \multicolumn{1}{r}{$\pm$0.25} & \multicolumn{1}{r}{$\pm$0.62} & \multicolumn{1}{r}{$\pm$0.52} &\multicolumn{1}{r}{$\pm$0.46}      \\ \bottomrule
\end{tabular}
\vspace{-0.4cm}
\end{table}

%% file: figures/6_loss_vs_fgsm_pgd_eps.tex
\begin{figure}[t]
        \centering
        {\includegraphics[width=\linewidth]{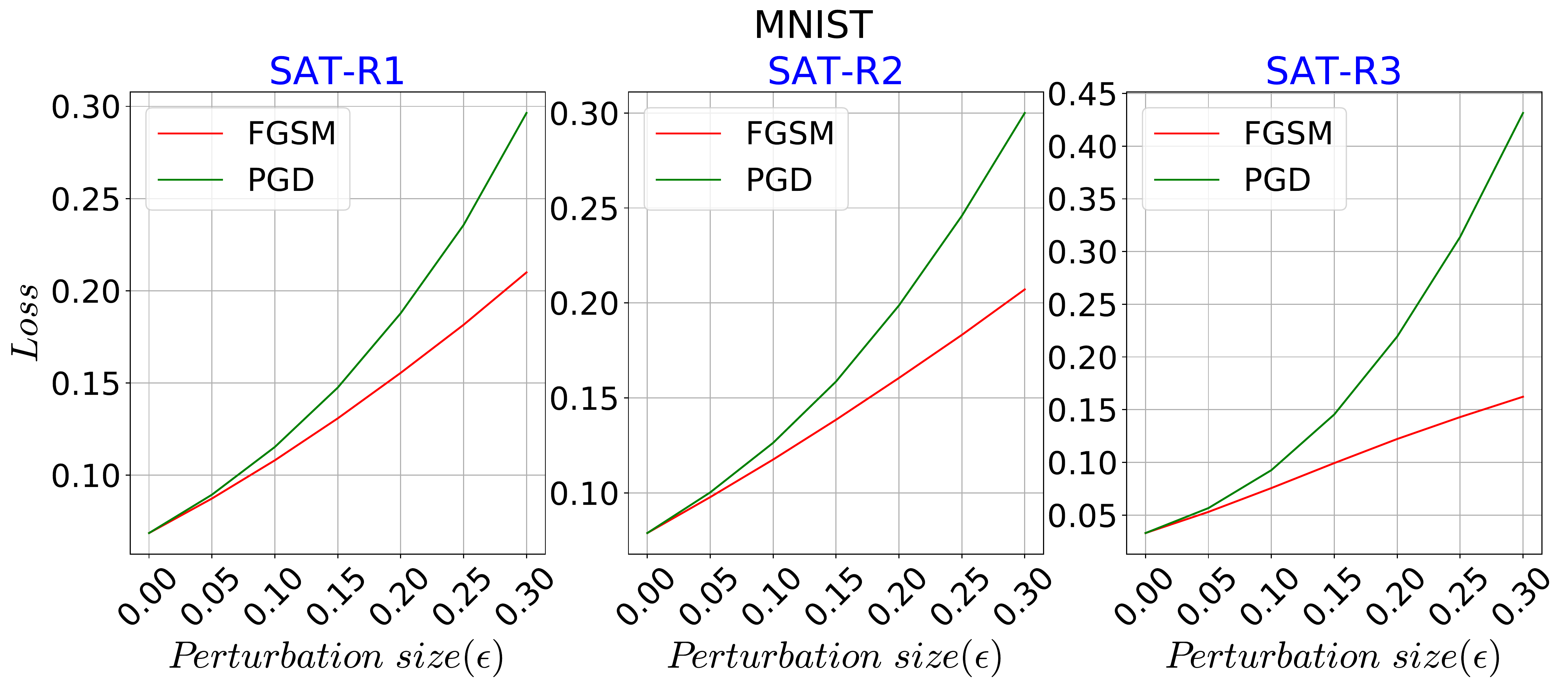}}
        {\includegraphics[width=\linewidth]{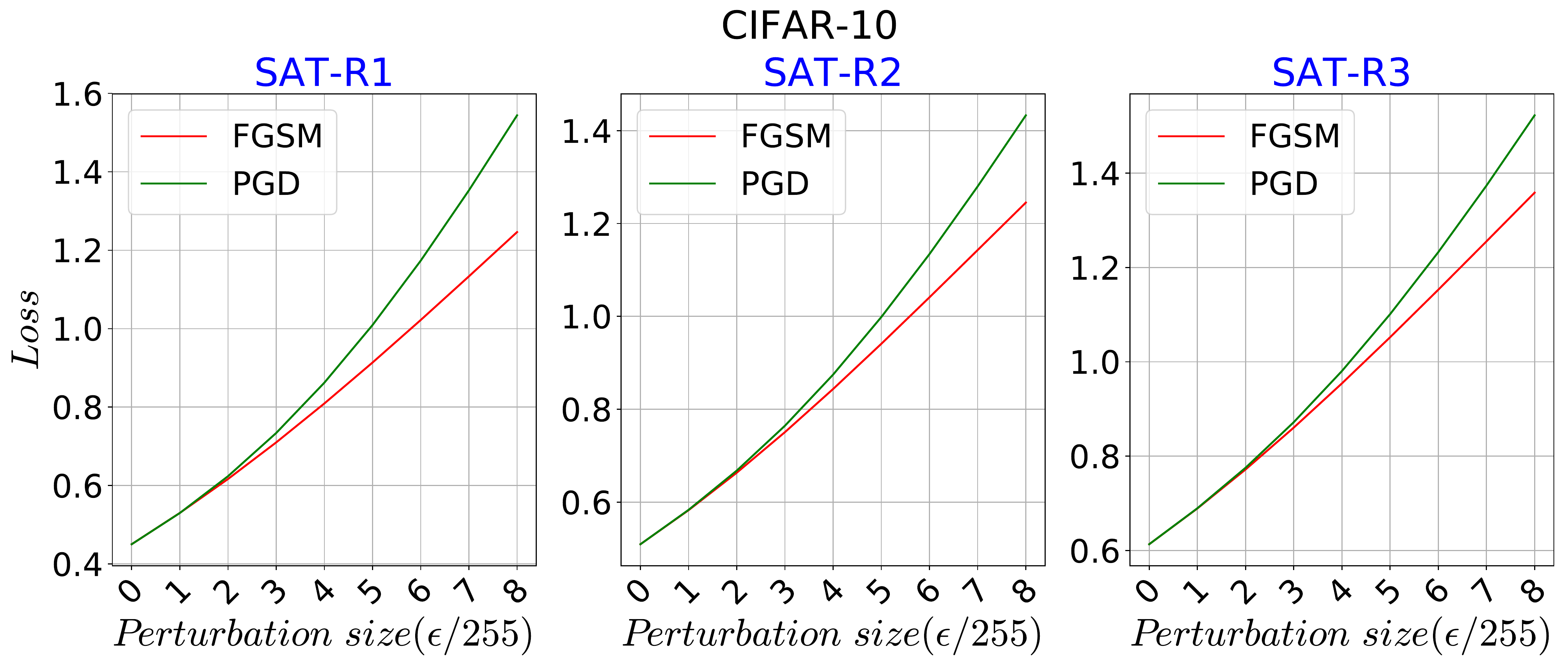}}
        {\includegraphics[width=\linewidth]{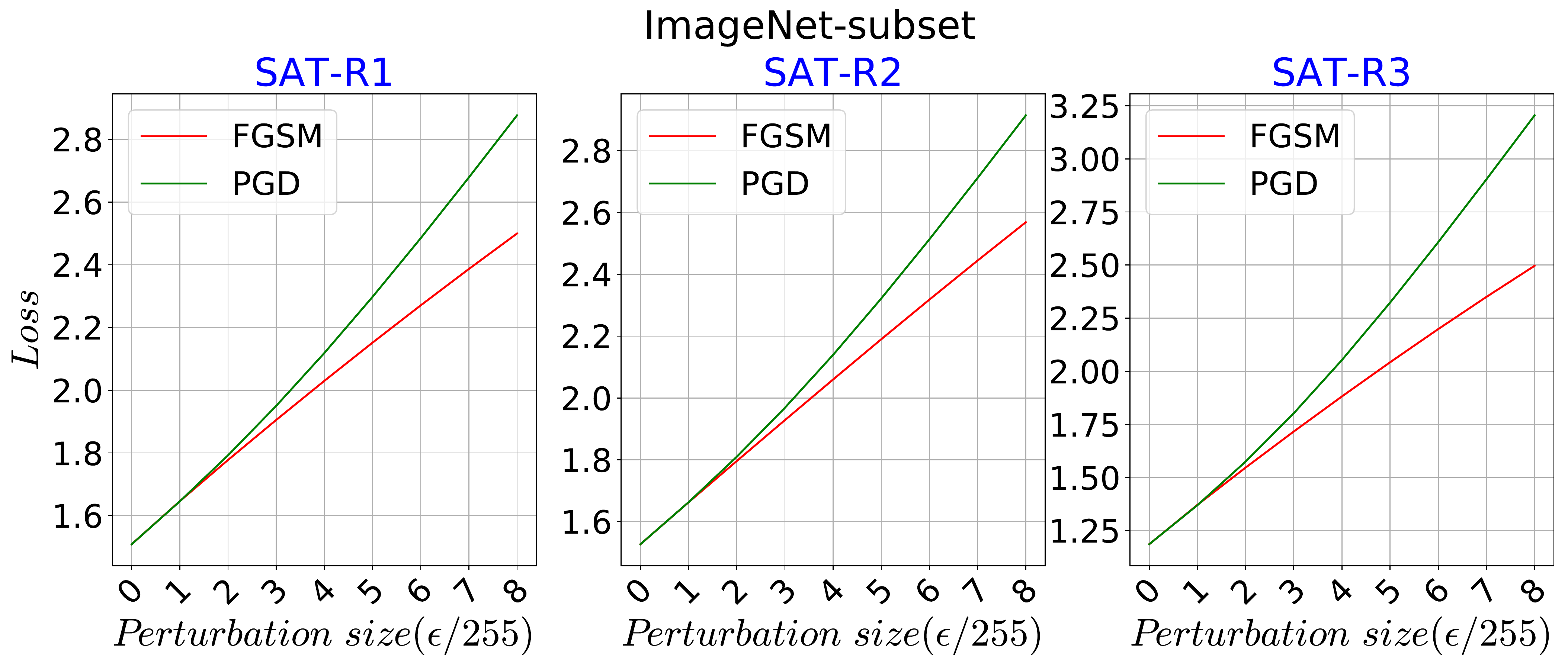}}
        \caption{Plot of the average loss on the test set versus perturbation size ($\epsilon$) of FGSM and PGD attacks, obtained for models trained using the proposed regularizers. Columns represent the training method and rows represent the dataset. Row-1: MNIST, row-2: CIFAR-10, and row-3: ImageNet-subset. Column-1: SAT-R1, column-2: SAT-R2, and column-3: SAT-R3.}
        \label{figure:proposed_loss_vs_fgsm_pgd_eps}
\end{figure}

%% file: figures/7_proposed_decision_surface.tex
\begin{figure}[h!]
        \centering
        {\includegraphics[width=0.82\linewidth]{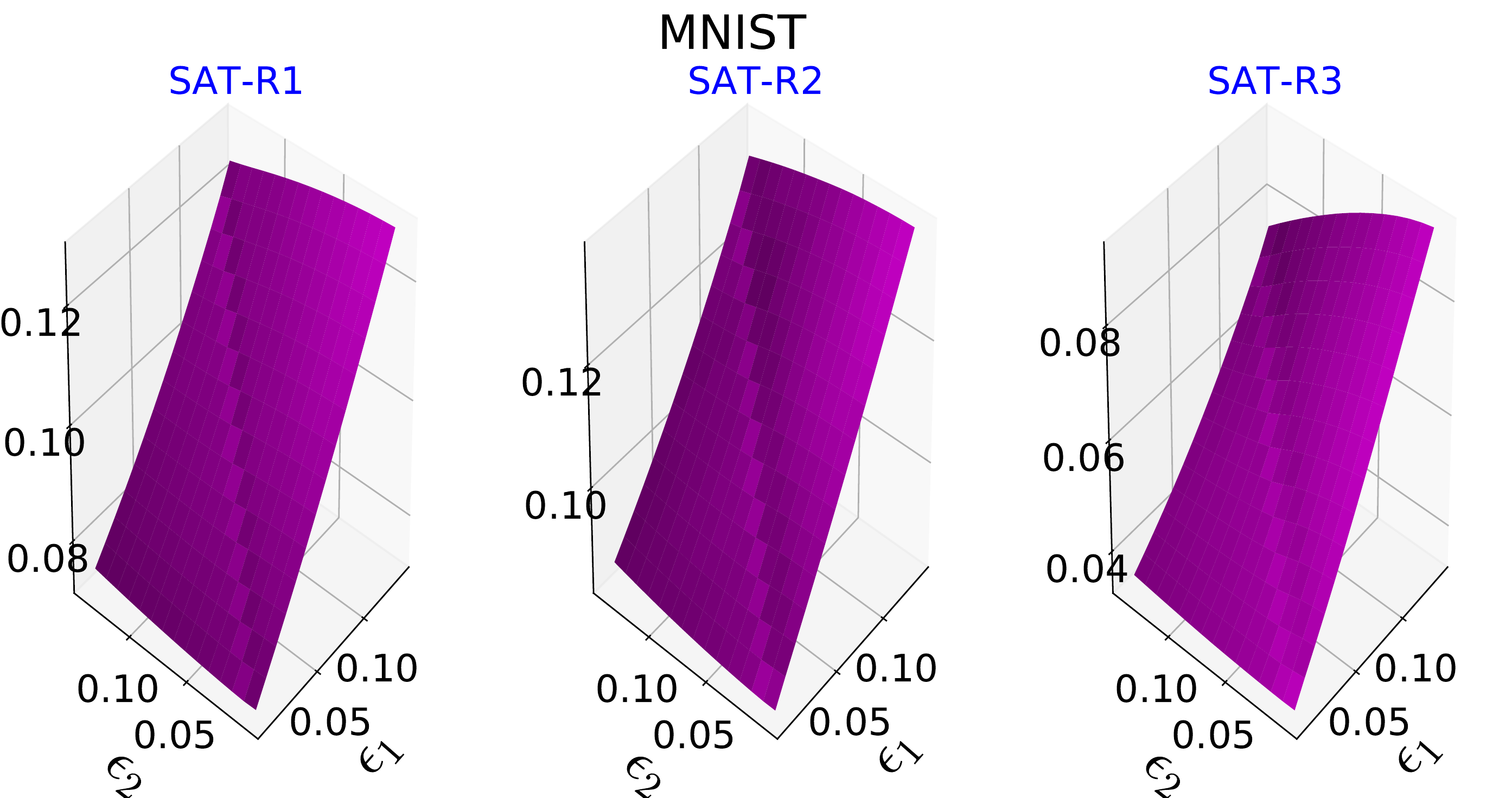}}
        {\includegraphics[width=0.82\linewidth]{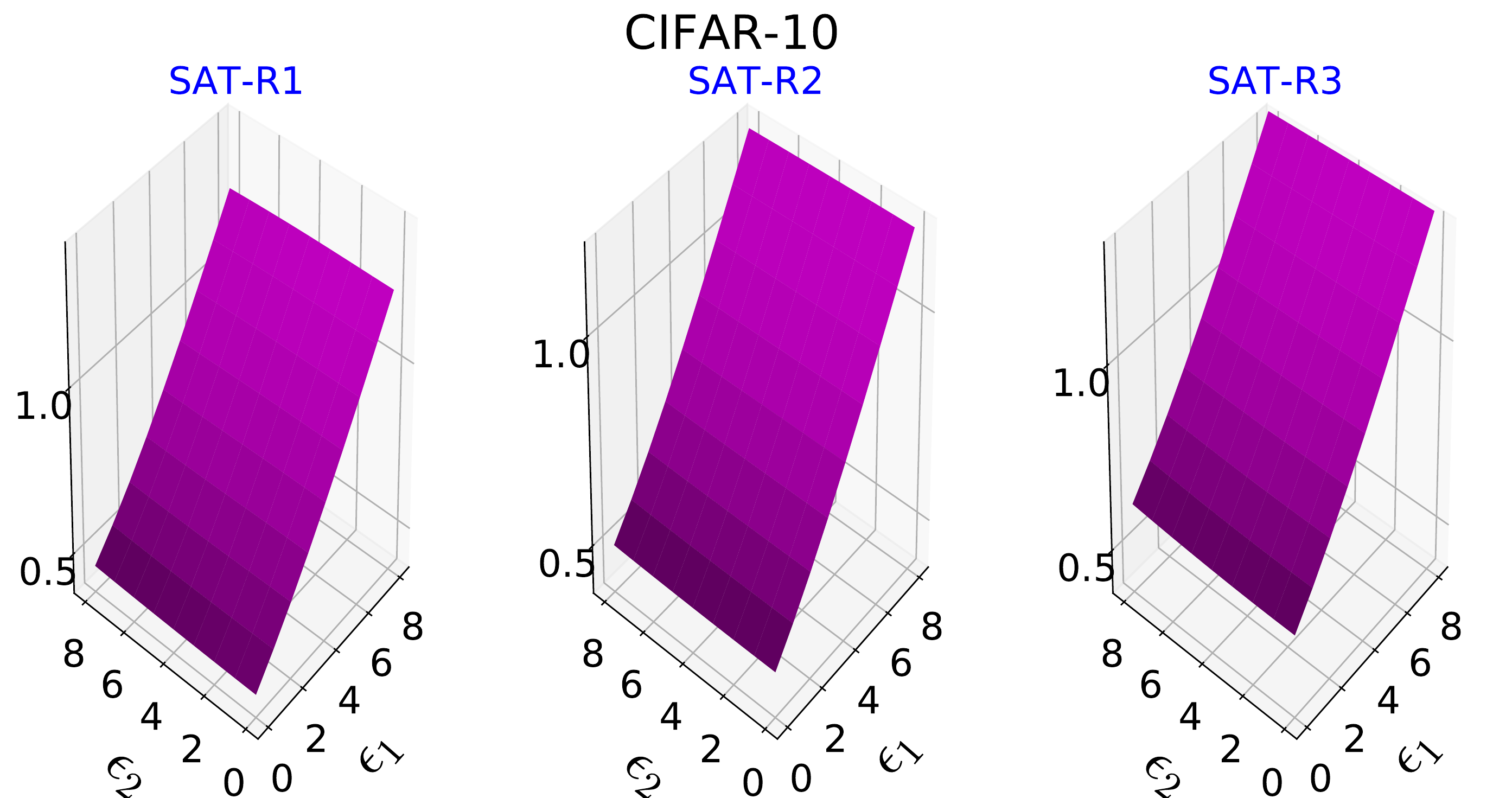}}
        {\includegraphics[width=0.82\linewidth]{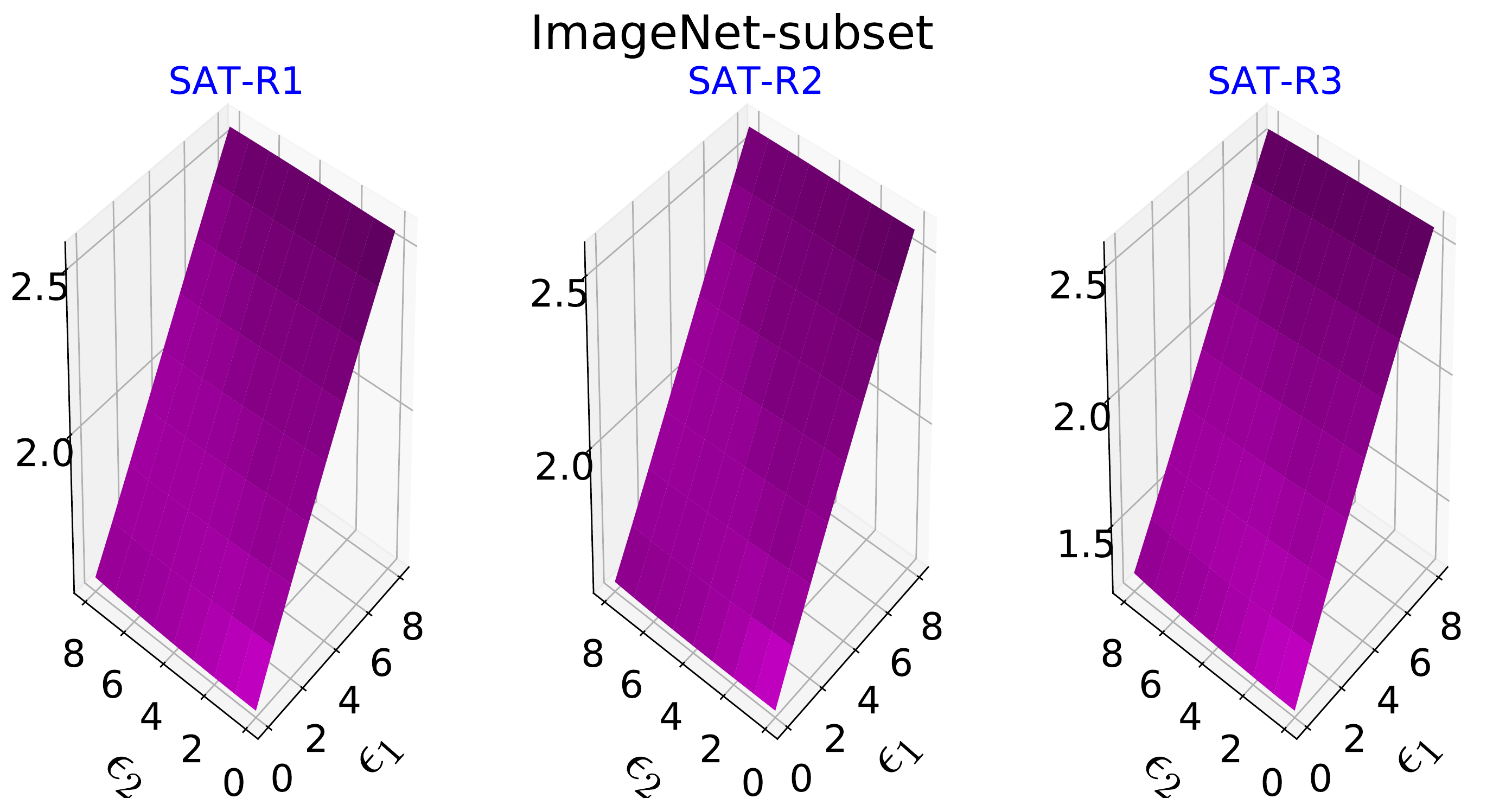}}
        \caption{Loss surface plots obtained for models trained using the proposed regularizers. Note that, no sharp curvature can be observed on the loss surface. Columns represent the training method and rows represent the dataset. Row-1: MNIST, row-2: CIFAR-10, and row-3: ImageNet-subset. Column-1: SAT-R1, column-2: SAT-R2, and column-3: SAT-R3.}
        \label{figure:proposed_decision_surface}
\end{figure}